\documentclass[10pt,twocolumn,letterpaper]{article}
\usepackage{iccv}
\usepackage{times}
\usepackage{epsfig}
\usepackage{graphicx}
\usepackage{amsmath}
\usepackage{amssymb}
\usepackage{breqn}
\usepackage{tabularx}
\usepackage{multirow}
\usepackage{color}
\graphicspath{{figures/}}
\usepackage{lineno}
\usepackage{subfigure}
\usepackage{epstopdf}
\usepackage{algorithm}
\usepackage{algpseudocode}
\usepackage{amsmath}
\usepackage[numbers,sort]{natbib}
\newcommand{\tabincell}[2]{\begin{tabular}{@{}#1@{}}#2\end{tabular}}

\usepackage[pagebackref=true,breaklinks=true,letterpaper=true,colorlinks,bookmarks=false]{hyperref}

\iccvfinalcopy 

\def\iccvPaperID{2076} 

\ificcvfinal\fi
\begin{document}

\title{Data Augmentation for Object Detection via Progressive and Selective Instance-Switching}

\author{Hao Wang$^{1}$, Qilong Wang$^{2}$,  Fan Yang$^{1}$, Weiqi Zhang $^1$, Wangmeng Zuo$^{1,}$\thanks{Corresponding author.}\\
$^1$Harbin Institute of Technology, China{},
$^2$Tianjin University, China\\
{\tt\small ddsywh@yeah.net, qlwang@tju.edu.cn, buddhisant@outlook.com }\\
{\tt\small amote530@163.com, wmzuo@hit.edu.cn}
}


\maketitle

\begin{abstract}
  Collection of massive well-annotated samples is effective in improving object detection performance but is extremely laborious and costly. Instead of data collection and annotation, the recently proposed Cut-Paste methods ~\cite{dwibedi2017cut,dvornik2018modeling} show the potential to augment training dataset by cutting foreground objects and pasting them on proper new backgrounds. However, existing Cut-Paste methods cannot guarantee synthetic images always precisely model visual context, and all of them require external datasets. To handle above issues, this paper proposes a simple yet effective instance-switching (IS) strategy, which generates new training data by switching instances of same class from different images. Our IS naturally preserves contextual coherence in the original images while requiring no external dataset. For guiding our IS to obtain better object performance, we explore issues of instance imbalance and class importance in datasets, which frequently occur and bring adverse effect on detection performance. To this end, we propose a novel Progressive and Selective Instance-Switching (PSIS) method to augment training data for object detection. The proposed PSIS enhances instance balance by combining selective re-sampling with a class-balanced loss, and considers class importance by progressively augmenting training dataset guided by detection performance. The experiments are conducted on the challenging MS COCO benchmark, and results demonstrate our PSIS brings clear improvement over various state-of-the-art detectors (e.g., Faster R-CNN, FPN, Mask R-CNN and SNIPER), showing the superiority and generality of our PSIS. Code and models are available at: \url{ https://github.com/Hwang64/PSIS}.
\end{abstract}

\section{Introduction}
Object detection as one of core issues in computer vision community has been covered with many applications, e.g., face recognition~\cite{ranjan2019hyperface}, robotic grasping~\cite{kumra2017robotic} and human interaction~\cite{gkioxari2018detecting}. In recent years, object detection greatly benefits from the rapid development of deep convolutional neural networks (CNNs)~\cite{simonyan2014very,he2016deep,szegedy2017inception,huang2017densely}, whose performance is heavily dependent on a mass of labeled training images. However, a large-scale well-annotated dataset for object detection is much more laborious and costly, compared with annotation of the whole images for classification. One potential way to efficiently acquire more training images is augmentation of the existing datasets or generation of synthetic images in an effective way.

In terms of data augmentation for object detection, traditional methods perform geometrical transformations (e.g., horizontal flipping~\cite{ren2015faster}, multi-scale strategy~\cite{singh2018sniper}, patch crop~\cite{liu2016ssd} and random erasing~\cite{zhong2017random}) on the original images to vary their spatial structures. However, these methods hardly change the visual content and context of objects, bringing little increase in the diversity of training dataset. Recently, some researches~\cite{gupta2016synthetic,dwibedi2017cut,georgakis2017synthesizing,dvornik2018modeling,lee2018context} show generation of new synthetic images in \emph{Cut-Paste} manner is a potential way to augment detection datasets. Specifically, given the cutting foreground objects, these methods paste them on some background images, where the pasted positions are estimated by considering 3D scene geometry~\cite{gupta2016synthetic}, in a random manner~\cite{dwibedi2017cut}, by predicting support surfaces~\cite{georgakis2017synthesizing} or by modeling visual context~\cite{dvornik2018modeling,lee2018context}. Meanwhile, above Cut-Paste methods~\cite{dvornik2018modeling,lee2018context} show appropriate visual context plays a key role in data augmentation for object detection. However, it is very difficult for the existing Cut-Paste methods to precisely model visual context for the whole time, although some prediction mechanisms are carefully designed. Furthermore, they require external datasets for collecting foreground/background images or both of them.

\begin{figure}[!t]
	\begin{center}
        \includegraphics[width=1.0\linewidth]{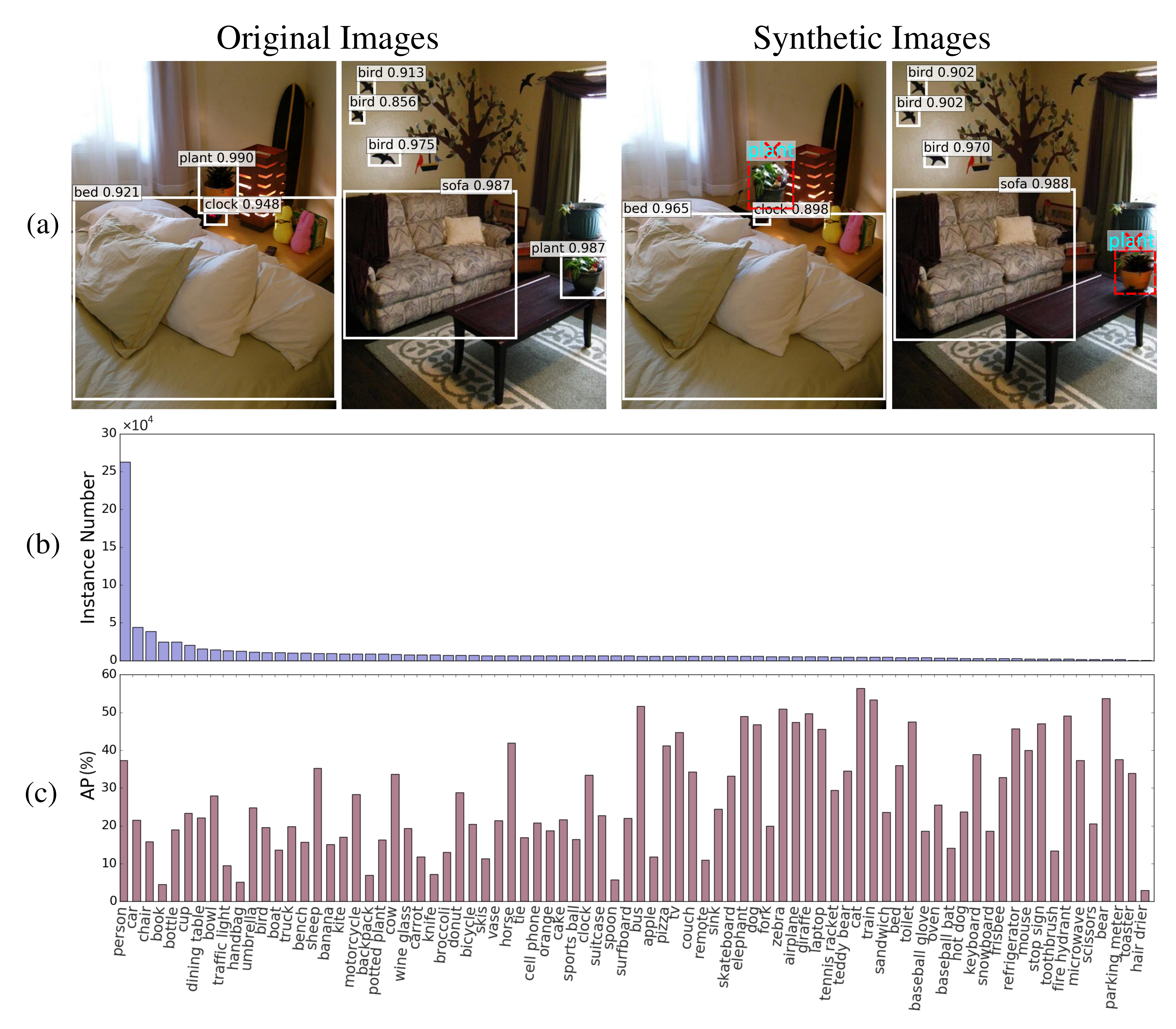}\\
		\caption{Some analysis based on Faster R-CNN~\cite{ren2015faster} with ResNet-101~\cite{he2016deep} on MS COCO~\cite{lin2014microsoft}. (a) Comparing a pair of the original images with a pair of synthetic ones generated by our instance-switching (IS) strategy. Clearly, the synthetic images naturally preserve  contextual coherence in the original ones. Meanwhile, the detector can correctly detect objects  in the original images, but miss the switched instances (i.e., plant, denoted by red dashed boxes) in synthetic images. (b) shows distribution of instances in each class and (c) gives average precision (AP) of each class.}
		\label{motivation}
	\end{center}
\end{figure}

To overcome above issues, this paper proposes a simple yet efficient instance-switching (IS) strategy. Specifically, given a pair of training images containing instances of same class, our IS method generates a new pair of training images by switching the instances of same class by taking shape and scale information of instances into account. As illustrated in  Figure~\ref{motivation} (a), our IS strategy can always preserve contextual coherence in the original images. Meanwhile, a detector (i.e., Faster R-CNN~\cite{ren2015faster} with ResNet-101~\cite{he2016deep}) trained on original dataset can correctly detect objects in the original images, but miss the switched instances (i.e., plant, denoted by red dashed boxes) in synthetic images, indicating IS has ability to increase the diversity of samples. In addition, our IS clearly requires no external datasets, and naturally keeps more coherence in visual context, compared with methods those paste the cutting foreground objects on a new background~\cite{gupta2016synthetic,dwibedi2017cut,dvornik2018modeling}.

The attendant problem is \emph{what criteria do we follow to perform IS?} We first make analysis using Faster R-CNN with ResNet-101 on the most widely used MS COCO~\cite{lin2014microsoft}. Figure~\ref{motivation} (b) and (c) show distribution of instances in each class and average precision (AP) of each class, respectively. We can clearly see that distribution of instances is long-tailed and detection performance of each class is quite different. Many works demonstrate data imbalance brings side effect on image classification ~\cite{cheng2018revisiting,dong2018imbalanced,huang2016learning,khan2018cost,cui2019class,wang2019dynamic}. Meanwhile, detection difficulties of different classes greatly vary, and distribution of instances and detection performance of each class do not always behave in the same way. Above issues of instance imbalance and class importance frequently occur in real-world applications and inevitably bring adverse effect on detection performance. However, few object detection method is concerned with these issues and the simple IS also lacks the ability to handle issue of instance imbalance and consider class importance in datasets. To this end, we explore a progressive and selective scheme for our IS strategy to generate synthetic images for object detection while training detectors with a class-balanced loss. The resulted method is called Progressive and Selective Instance-Switching (PSIS), which enhances instance balance by selectively performing IS (i.e., inversely proportional to the original instance frequency) for adjusting distribution of instances~\cite{chawla2002smote,he2008learning,buda2018systematic} and combining with a class-balanced loss~\cite{cui2019class}. To consider class importance, PSIS augments training dataset in a progressive manner, which is guided by detection performance (i.e., increase of times of IS for classes with lowest APs). The experiments are conducted on MS COCO dataset ~\cite{lin2014microsoft} to evaluate out PSIS.

The contributions of this paper are summarized as follows: (1) We propose a simple yet efficient instance-switching (IS) strategy to augment training data for object detection. The proposed IS can increase the diversity of samples, while preserving contextual coherence in the original images and requiring no external datasets. (2)  We propose a novel Progressive and Selective Instance-Switching (PSIS) method to guide our IS strategy can enhance instance balance and consider class importance, which are ubiquity but few detection methods pay attention to. Our PSIS can further improve detection performance. (3) We thoroughly evaluate the proposed PSIS on challenging MS COCO benchmark, and experimental results show our PSIS is superior and complementary to existing data augmentation methods and brings clear improvement over various state-of-the-art detectors (e.g., Faster R-CNN~\cite{ren2015faster}, Mask R-CNN~\cite{he2017mask} and SNIPER~\cite{singh2018sniper}), while our PSIS has the potential to improve performance of instance segmentation~\cite{he2017mask}.

\begin{figure*}[!t]
	\begin{center}
		\includegraphics[width=0.9\linewidth]{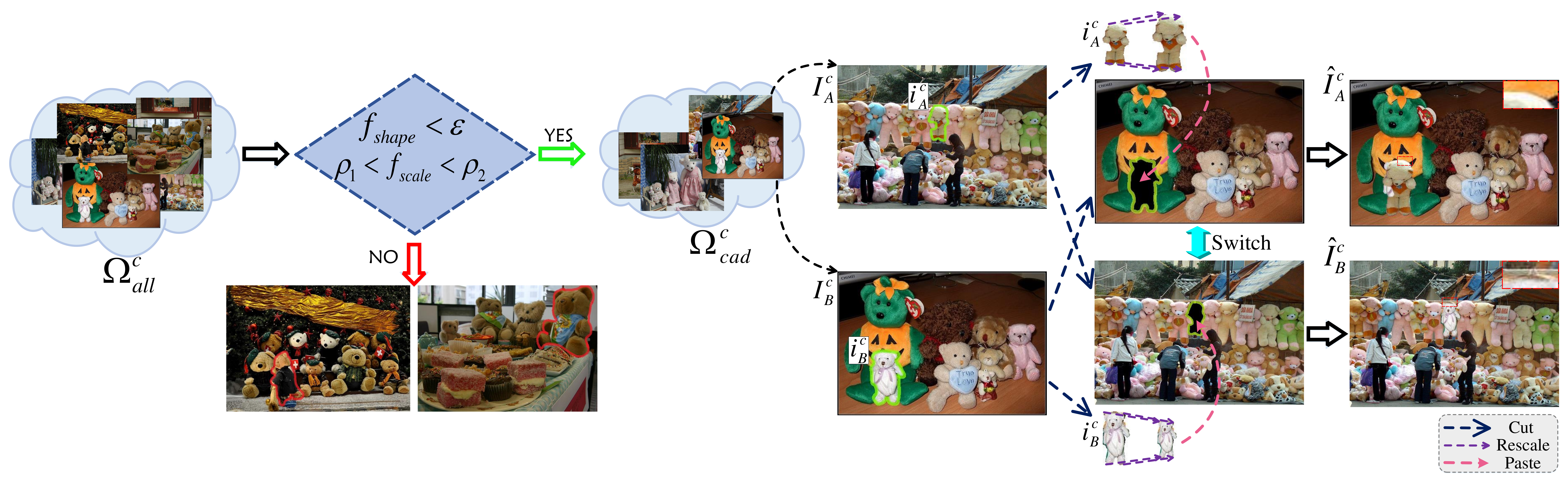}\\
		\caption{Overview of instance-switching strategy for image generation. We first select a candidate set $\Omega^{c}_{cad}$ from all training data $\Omega^{c}_{all}$ for each class based on shape and scale of instance. For a quadruple $\{I_{A}^{c}$,$I_{B}^{c},i_{A}^{c},i_{B}^{c}\}$ in $\Omega^{c}_{cad}$, we switch the instances $i_{A}^{c}$ and $i_{B}^{c}$ through rescaling and cut-paste. Finally, Gaussian blurring is used to smooth the boundary artifacts. Thus, synthetic images $\{\hat{I}_{A}^{c}$,$\hat{I}_{B}^{c}\}$ are generated. }\label{fig1}
	\end{center}
\end{figure*}

\section{Related Work}
Our work focuses on data augmentation for object detection. Albeit many data augmentation methods are proposed for deep CNNs in image classification task~\cite{devries2017dataset,inoue2018data,takahashi2018data,zhang2017mixup,devries2017improved}, it is not clear whether they are suitable for object detection or not. The traditional data augmentation for object detection performs geometrical transformations on  images~\cite{ren2015faster,liu2016ssd,fu2017dssd,he2017mask}, and some works use random occlusion mask to dropout some part of images~\cite{zhong2017random} or learn occlusion mask on convolutional feature maps~\cite{0004SG17} for generating hard positive samples. An alternative solution is to generate new synthetic images in \emph{Cut-Paste} manner~\cite{gupta2016synthetic,dwibedi2017cut,georgakis2017synthesizing,dvornik2018modeling,lee2018context}. Compared with above methods, our method can better increase the diversity of samples than traditional ones, and model more precise visual context than Cut-Paste methods in an efficient way. Generative Adversarial Networks (GANs)~\cite{goodfellow2014generative,ChenCDHSSA16,OordKK16,ArjovskyCB17} have shown promising ability for image synthesis. However, these methods are unable to automatically annotate bounding boxes of objects, so that they cannot be directly used for object detection. The recently proposed compositional GAN~\cite{azadi2018compositional} introduces a self-consistent composition-decomposition network to generate images by combining two instances with explicitly learning possible interactions, but it is incapable of generating images with complex context. Remea \etal ~\cite{remez2018learning} use a copy-and-paste GAN to generate images for weakly segmentation tasks. However, synthetic images generated by GANs usually suffer from a clear domain gap with the real images, bringing the side effect for subsequent applications~\cite{ShrivastavaPTSW17}.

The way of our PSIS method to enhance instance balance is related to works those handle the issue of class imbalance in image classification~\cite{cui2019class,dong2018imbalanced,wang2019dynamic,huang2016learning,chawla2002smote,he2008learning}. Among them, methods  \cite{chawla2002smote,he2008learning,buda2018systematic} balance distribution of classes by re-sampling for those containing less training samples. Another kind of methods perform sample re-weighting by introducing some cost-sensitive losses~\cite{huang2016learning,khan2018cost,sarafianos2018deep,cui2019class}, which can alleviate the over-fitting caused by duplicated samples in re-sampling process. Wang \etal ~\cite{wang2019dynamic} propose an on-line method to adaptively fuse the sampling strategy with loss learning using two level curriculum schedulers.  But differently, our PSIS focuses on balance of instances rather than images, as one image may contain multiple instances. Moreover, we combine re-sampling with sample re-weighting for better enhancing instance balance. Additionally, Focal loss~\cite{lin2017focal} is proposed to handle issue of class imbalance in one-stage detectors, but it only focuses on two classes (i.e., foreground and background). Our PSIS employs a progressive augmentation process, which is related to \cite{cheng2018revisiting}. The latter proposes a Drop and Pick training policy to select samples for reducing training computation in image classification task. In contrast, our PSIS progressively increases number of instances for classes with lowest APs.

\section{Proposed Method}
In this section, we introduce the proposed Progressive and Selective Instance-Switching (PSIS) method. We first describe our Instance-Switching (IS) strategy for synthetic images generation. Then, selective re-sampling and class-balanced loss are introduced to enhance instance balance. Additionally, we perform IS in a progressive manner for considering class importance. Finally, we show the overview of our PSIS for object detection.

\subsection{Instance-Switching for Image Generation}~\label{sec3.1}
Previous works \cite{dvornik2018modeling,lee2018context} show appropriate visual context plays a key role in synthetic images for object detection. To precisely model visual context of synthetic images while increasing the diversity of training samples, this paper proposes an instance-switching (IS) strategy for synthetic image generation. Figure~\ref{fig1} illustrates the core idea of our IS for image generation. Let $\Omega^{c}_{all}$ be the training set of class $c$ (e.g., teddy bear) involving $N$ images. To guarantee visual context of synthetic images as coherence as possible, we define a candidate set (indicted by $\Omega^{c}_{cad}$) based on segmentation masks of instances\footnote{In this paper, our experiments are conducted on MS COCO dataset~\cite{lin2014microsoft}, where segmentation mask of each instance can be reliably obtained using \href{https://github.com/cocodataset/cocoapi}{COCO API}. Note that all existing Cut-Paste based data augmentation methods for object detection employ segmented objects or masks~\cite{gupta2016synthetic,dwibedi2017cut,georgakis2017synthesizing,dvornik2018modeling}.}.

Specifically, our $\Omega^{c}_{cad}$ consists of a set of quadruples $\{I_{A}^{c}$,$I_{B}^{c},i_{A}^{c},i_{B}^{c}\}$, where $I_{A}^{c}$, $I_{B}^{c}$, $i_{A}^{c}$ and $i_{B}^{c}$ satisfy the following conditions:
\begin{align}\label{IS_set}
&\Omega^{c}_{cad}:\{I_{A}^{c},I_{B}^{c}\}, \,\, I_{A}^{c}, I_{B}^{c}\in\Omega^{c}_{all}, \,\, i_{A}^{c}\in I_{A}^{c},\,\, i_{B}^{c}\in I_{B}^{c},\\
&s.t., \,\,\, f_{shape}(i_{A}^{c}, i_{B}^{c}) < \varepsilon, \,\,\, \rho_{1} < f_{scale}(i_{A}^{c}, i_{B}^{c}) < \rho_{2},  \nonumber
\end{align}
where  $I_{A}^{c}$ and $I_{B}^{c}$ are two images in $\Omega^{c}_{all}$; $i_{A}^{c}$ and $i_{B}^{c}$ are instances of label $c$ in images $I_{A}^{c}$ and $I_{B}^{c}$, respectively. $f_{shape}$ and $f_{scale}$ are two functions matching shapes and scales of instances $i_{A}^{c}$ and $i_{B}^{c}$. Let $m_{A}^{c}$ and $m_{B}^{c}$ be masks (binary images) of $i_{A}^{c}$ and $i_{B}^{c}$, we align them and rescale them to the same size, then the normalized masks are denoted as $\widehat{m}_{A}^{c}$ and $\widehat{m}_{B}^{c}$. Accordingly, the function $f_{shape}$ can be computed based on sum square difference (ssd)
~\cite{zhu2015learning,lalonde2007using}, i.e.,
\begin{align}\label{f_shape}
f_{shape}(i_{A}^{c}, i_{B}^{c}) = \frac{\text{ssd}(\widehat{m}_{A}^{c},\widehat{m}_{B}^{c})}{max(area(\widehat{m}_{A}^{c}),area(\widehat{m}_{B}^{c}))},
\end{align}
where $area(Q)$ indicates area of $Q$, counted by the number of one in $Q$. $max$ is a maximum function. Then, the function $f_{scale}$ has
\begin{align}\label{f_scale}
f_{scale}(i_{A}^{c}, i_{B}^{c}) = area(m_{B}^{c})/area(m_{A}^{c}).
\end{align}

According to Eqns.~(\ref{IS_set}), (\ref{f_shape}) and (\ref{f_scale}), $\Omega^{c}_{cad}$ consists of a set of quadruples $\{I_{A}^{c}$,$I_{B}^{c},i_{A}^{c},i_{B}^{c}\}$, where instances $i_{A}^{c}$ and $i_{B}^{c}$ have similar shapes (i.e., shape differences are less than threshold $\varepsilon$) and are in controlled scaling ratios (i.e., scaling ratios  range from $\rho_{1}$ to $\rho_{2}$). Given the candidate set $\Omega^{c}_{cad}$, we can generate new training images $\{\hat{I}_{A}^{c}$,$\hat{I}_{B}^{c}\}$ by switching instances $i_{A}^{c}$ and $i_{B}^{c}$ through rescaling and cut-paste. Finally, we follow~\cite{dwibedi2017cut} to employ Gaussian blurring for smoothing the boundary artifacts. Once the quadruple $\{I_{A}^{c}$,$I_{B}^{c},i_{A}^{c},i_{B}^{c}\}$ completes IS, we will remove them from $\Omega^{c}_{cad}$ to avoid duplicated sampling. Clearly, our IS method can preserve contextual coherence in the original images and requires no external datasets.

\subsection{Enhancement of Instance Balance}~\label{sec3.2}
The data in many real-world applications are long-tailed distributions. As shown in Figure~\ref{motivation}, the most widely used object detection dataset~\cite{lin2014microsoft} also suffers from extreme imbalance distribution of instances, bringing the challenge for detection methods. To alleviate this issue, we propose to exploit a selective re-sampling strategy for balancing distribution of instances and a class-balanced loss~\cite{cui2019class} for avoiding over-fitting of re-sampling.

\subsubsection{Selective Re-sampling Strategy}~\label{sec3.2.1}
\begin{figure}[!t]
	\begin{center}
		\includegraphics[width=0.86\linewidth]{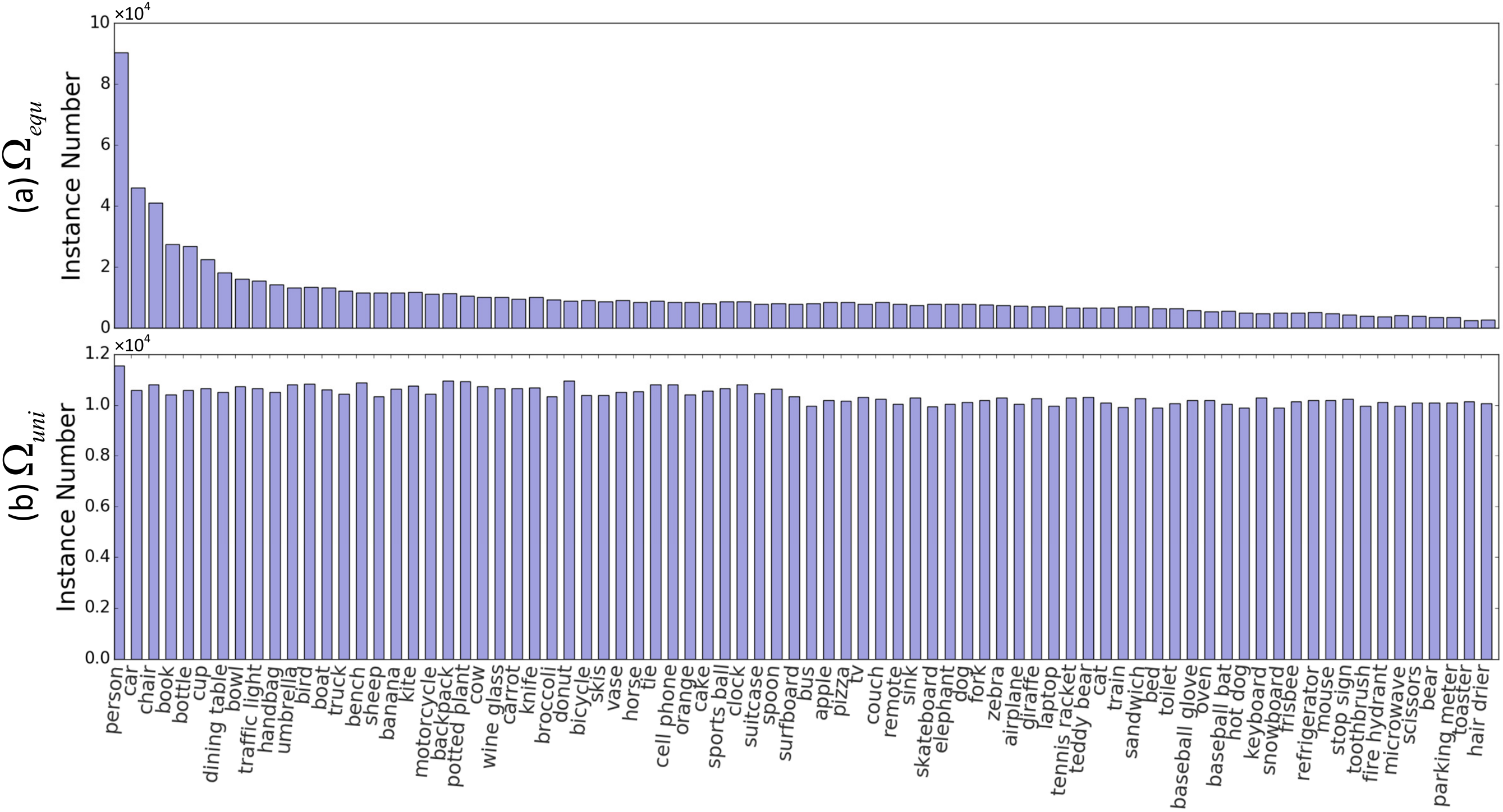}
		\caption{Distributions of instances in equally sampled dataset $\Omega_{equ}$ (a) and adjustment dataset $\Omega_{uni}$ (b). Clearly, our re-sampling method makes distribution of instances more uniform, leading to better detection performance (refer to Section~\ref{sec4.1.2}).}
		\label{rda}
	\end{center}
\end{figure}

To obtain an instance-balanced training dataset, a natural scheme is to select the same number of quadruples in each $\Omega^{c}_{cad}$, and perform IS to generate synthetic images. The resulted dataset is dubbed by $\Omega_{equ}$, and the selected set of quadruples in $c$-th class is indicated by $\Omega_{equ}^{c}$. Comparing Figure~\ref{motivation} (b) with  Figure~\ref{rda} (a), distribution of instances in $\Omega_{equ}$ is more balanced than one in original training dataset $\Omega_{ori}$. However, one image usually contains different instances of various numbers, so distribution of instances in the equally sampled dataset $\Omega_{equ}$ is still far away from a uniform one. Hence, we propose a selective re-sampling strategy to further adjust distribution of instances in $\Omega_{equ}$.

Let small and large classes indicate the classes with fewer and more instances, respectively. Our selective re-sampling strategy is developed based on a straightforward rule: (1) On the one hand, we pick more images for small classes, where each picked image should contain instances of small classes as many as possible while involving instances of large classes as few as possible; (2) On the other hand, we drop images for large classes, where the dropped images have opposite situation to the picked ones. Intuitively, if we persistently carry out above drop-pick process, distribution of instances in $\Omega_{equ}$ will tend to be uniform.

In our case, we drop or pick the image $I$ if it satisfies the following conditions:
\begin{eqnarray}\label{Eq_SR1}
\frac{M_{c}-M_{I_{c}}}{M_{equ}-M_{I}} < \frac{M_{c}}{M_{equ}} < \frac{M_{c}+M_{I_{c}}}{M_{equ}+M_{I}}
\end{eqnarray}
where $M_{c}$ and $M_{equ}$ are numbers of instances in $\Omega_{equ}^{c}$ and  $\Omega_{equ}$, respectively; $M_{I_{c}}$ and $M_{I}$ indicate number of instances of $c$-th class and number of all instances in $I$, respectively. Furthermore, it is difficult to directly handle all classes as number of classes increases. So we carry out above drop-pick within two stages. In the first stage, we handle maximum and minimum classes, and the drop-pick process stops when percentage of instances in minimum and maximum classes are 1.5 times and half of original ones. Once drop-pick process stops, maximum and minimum classes will be changed, so we iteratively carry out this process 20 times. In the second stage, we extend above process to all classes. After our selective re-sampling strategy, as shown in Figure~\ref{rda} (b), distribution of instances in $\Omega_{equ}$ is adjusted to be approximately uniform, which is indicated by $\Omega_{uni}$.

\subsubsection{Class-Balanced Loss}~\label{sec3.2.2}
Above selective re-sampling strategy can generate a synthetic training dataset where distribution of instances is approximately uniform, but it is still very difficult to make the whole training datasets (i.e., $\Omega_{uni}+\Omega_{ori}$) obey a uniform distribution. Moreover, over-sampling easily generates the duplicated images, leading to over-fitting~\cite{cui2019class}. The similar phenomenon is also found in our experiments (refer to Section~\ref{sec4.1.2} for details). To this end, we exploit the recently proposed class-balanced loss~\cite{cui2019class} to re-weight instances, which reformulates the softmax loss by introducing a weight that is inversely proportional to the number of instances in each class. Given an instance with label $c$ and the prediction vector $\mathbf{z}=[z_1,z_2,\dots,z_C]^{\top}$, where $C$ is the number of classes.  The class-balanced loss can be computed as:
\begin{equation}\label{cbl}
\ell_{CB}(\mathbf{z},c)=-\frac{\gamma}{1-(1-\gamma)^{n_{c}}}\log(\frac{\exp(z_c)}{\sum_{j=1}^{C}\exp(z_j)})
\end{equation}
where $n_{c}$ is the number of instances belonging to the class of $c$, and $0< \gamma \leqslant 1$ is a regularized constant to control the effect of number of instances on final prediction. In particular, Eqn.~(\ref{cbl}) will degenerate to the standard softmax loss if $\gamma=1$, and $\gamma \rightarrow 0$ will enlarge the effect of number of instances, which is evaluated in Section~\ref{sec4.1.2}.

\subsection{Progressive Instance-Switching}~\label{sec3.3}
In above section, we introduce two strategies to enhance instance balance. As shown in Figure~\ref{motivation} (b) and (c), detection performance (i.e., AP) of each class is quite different, while AP of each class does not always show positive correlation with distribution of instances.  Above observations encourage us to pay more attention on these classes with lowest APs for further improving detection performance. To this end, we introduce a progressive instance-switching for data augmentation, which generates more training images for the classes with lowest APs in a self-supervised learning manner.

Specifically, we first train a detector using a predefined training dataset $\Omega_{train}$. After $T$ training epochs, we evaluate the current detector on the validation set and pick up $K$ classes with the lowest APs. In order to control the number of augmented images for avoiding over-fitting and breaking of balance, we perform selective instance-switching (Section~\ref{sec3.2.1}) to generate training images for $c$-th class ($c$ is one of $K$ classes of lowest APs) based on the proportion $\theta_c$:
\begin{equation}\label{PIS}
\theta_c=p+\frac{p}{K}*Rank(c),\,\, 1 \leqslant c \leqslant K,
\end{equation}
where  $K$ and $p$ are number of classes and image percentage to be augmented, which are decided by cross validation. $Rank(c)$ indicates rank of AP of $c$-th class sorted in a descending order, e.g, $Rank(c)=K$ if $c$-th class has the lowest AP. The new generated dataset is denoted by $\Omega_{aug}$, given it we go on training the current detector using union set of $\Omega_{train}$ and $\Omega_{aug}$. According to Eqn.~(\ref{PIS}), our progressive instance-switching method can augment more training images for the classes with lower APs in a controlled range, which takes class importance into consideration.

\begin{algorithm}[htb]
	\caption{PSIS for Object Detection}
	\label{alg1}
	\begin{algorithmic}[1]
		\Require
		Original training dataset $\Omega_{ori}$, number of training epochs $T_{total}$;
		\Ensure
		Augmented dataset $\Omega_{PSIS}$, object detector $\mathcal{D}$;
		\State Generate a uniform augmented dataset $\Omega_{uni}$ (Section~\ref{sec3.2.1});
		\State $\Omega_{train} \leftarrow \Omega_{ori} \cup \Omega_{uni}$;
		\For{ $t \in [1,\dots,T_{total}]$}
		\State Train detector $\mathcal{D}_{t}$ with loss $\ell_{CB}$~(\ref{cbl}) on $\Omega_{train}$;
		\If { $\mod(t,\,T) = 0$ }
		\State Generate a new dataset $\Omega_{aug}$ based on $\mathcal{D}_{t}$ (Section~\ref{sec3.3});
		\State $\Omega_{train} \leftarrow \Omega_{train} \cup \Omega_{aug}$;
		\EndIf
		\EndFor
		\State $\Omega_{PSIS}  \leftarrow \Omega_{train}$;
	\end{algorithmic}
\end{algorithm}

\subsection{PSIS for Object Detection}~\label{sec3.4}
So far, we have introduced our proposed progressive and selective instance-switching (PSIS) method. Finally, we show how to apply our PSIS for object detection, which is summarized in Algorithm~\ref{alg1}. Specifically, we first generate a uniform augmented dataset $\Omega_{uni}$ based on the original training dataset $\Omega_{ori}$ as descried in Section~\ref{sec3.2.1}. Then we combine $\Omega_{ori}$ with $\Omega_{uni}$ as initial training dataset $\Omega_{train}$, and employ $\Omega_{train}$ to train the detector $\mathcal{D}$ with the class balance loss (\ref{cbl}). After $T$ training epochs, we augment the $\Omega_{train}$ using the current detector $\mathcal{D}_{t}$ following the rules in Section~\ref{sec3.3}, and continue training of detector on the union set of $\Omega_{train}$ and $\Omega_{aug}$. Finally, our PSIS algorithm outputs a object detector $\mathcal{D}$ and an augmented dataset $\Omega_{PSIS}$. As shown in our experiments, the augmented dataset $\Omega_{PSIS}$ can be flexibly adopted to other detectors or tasks.

\begin{table*}[!t]
	\begin{center}
	 \footnotesize
	 \caption{Results (\%) of Faster R-CNN (ResNet-101) on validation set of MS COCO 2017 under different training configurations. }\label{abstudy}
		\begin{tabular}{l|c|ccc|ccc|ccc|ccc}
			\hline
			\multirow{2}{*}{ Training sets} & \multirow{2}{*}{Detector}  & \multicolumn{3}{c|}{ Avg.Precision, IOU:} & \multicolumn{3}{c|}{Avg.Precision, Area:} & \multicolumn{3}{c|}{ Avg.Recall, \#Det:} & \multicolumn{3}{c}{Avg.Recall, Area:}\\
			\cline{3-14}
			& & 0.5:0.95 & 0.50 & 0.75 &  Small & Med. & Large & 1 &   10 &  100 &  Small &  Med. &  Large\\
			\hline
			 $\Omega_{ori}$  & \multirow{5}{*}{\tabincell{c}{Faster R-CNN\\(ResNet-101)\\ \cite{ren2015faster}}}  &  27.3 &  48.6 &  27.5 &  8.8 & 30.4 & 43.1 & 25.8 & 38.5 & 39.3 & 16.3 & 44.5 &  60.2\\
			 $\Omega_{ori}$ +  $\Omega_{equ}$ &  & 28.4 & 49.1 & 29.1 &  8.9 & 32.1 & 44.8 & 26.7 &  39.4 &  40.5 &  16.8 &  45.8 & 62.0\\
			 $\Omega_{ori}$ +  $\Omega_{uni}$ &   & 28.7 & 49.3 & 29.8 & 9.2 & 32.6 & 45.2 & 27.2 & 39.7 & 41.0 & 17.1 & 46.3 & 62.7\\
			 $\Omega_{ori}$ +  $\Omega_{uni}$ + $\ell_{CB}$ &  & 29.0 & 49.7 & 30.2 & 9.4 & 33.0 & 46.0 & 27.4 & 40.1 & 41.5 & 17.5 & 47.0 & 63.2\\
			 $\Omega_{PSIS}$ + $\ell_{CB}$ &  & \textbf{29.7} & \textbf{50.6} & \textbf{30.5} & \textbf{10.3} & \textbf{33.4} &  \textbf{48.0} & \textbf{27.6} & \textbf{40.5} &  \textbf{41.6} & \textbf{18.6} & \textbf{47.6} & \textbf{64.6}\\
			 \hline
		\end{tabular}
	\end{center}\vspace{-.4cm}
\end{table*}

\section{Experiments}

In this section, we evaluate our PSIS method on the challenging MS COCO 2017~\cite{lin2014microsoft}, which is the most widely used benchmark for object detection,  containing about 118k training, 5k validation and 40k testing images collected from 80 classes. We first make ablation studies using Faster R-CNN~\cite{ren2015faster} with  ResNet-101~\cite{he2016deep}. For determining candidate sets for our instance-switching, we fix the parameters $\varepsilon$,  $\rho_{1}$ and $\rho_{2}$ in Eqn.~(\ref{IS_set}) to $0.3$, $1/3$ and $3$, respectively. We train Faster R-CNN (ResNet-101) within 14 epochs (i.e., $T_{total}=14$ in Algorithm~\ref{alg1}) and set $T$ to $6$. The initial learning rate is set to $10^{-3}$, and decreases with a factor of 10 after 11 training epochs. The horizontal flipping is used for data augmentation. After generating augmented dataset ($\Omega_{PSIS}$) by our PSIS and Faster R-CNN, we directly adopt $\Omega_{PSIS}$ to four state-of-the-art detectors (i.e.,  FPN~\cite{lin2017feature}, Mask R-CNN~\cite{he2017mask}, SNIPER~\cite{singh2018sniper} and BlitzNet~\cite{dvornik2017blitznet}) for comparing with the related works. Finally, we verify the generalization ability of $\Omega_{PSIS}$ on instance segmentation task. For all evaluated detectors, Faster R-CNN, SNIPER and BlitzNet are implemented using the source codes released by the respective authors. We employ the publicly available toolkit~\cite{mmdetection2018} to implement FPN~\cite{lin2017feature} and Mask R-CNN~\cite{he2017mask}. All programs run on a server equipped with four NVIDIA GTX 1080Ti GPUs. Following the standard evaluation metric (e.g., mean Average Precision (mAP) under $IoU=[0.5:0.95]$), we report results on validation set and  \href{https://competitions.codalab.org/competitions/5181}{$test$-$dev$ 2017} evaluation server for ablation studies and comparison, respectively.

\subsection{Ablation studies}~\label{sec4.1}

In this subsection, we assess the effects of different components on our PSIS, including instance-switching strategy, instance balance enhancement and progressive instance-switching. To this end, we train Faster R-CNN~\cite{ren2015faster} on various training datasets under the exactly same experimental settings, and report the results on the validation set.

\vspace{-.3cm}
\subsubsection{Instance-switching Strategy}~\label{sec4.1.1}\vspace{-.5cm}

Using our instance-switching strategy in the equally sampling manner, we can generate a synthetic dataset $\Omega_{equ}$, which shares the same size (i.e., 118k) with the original training dataset $\Omega_{ori}$.
As compared in the top two rows of Table~\ref{abstudy}, combination of $\Omega_{ori}$ with $\Omega_{equ}$ can achieve consistent improvements over single $\Omega_{ori}$ under all evaluation metrics. In particular, Faster R-CNN (ResNet-101) trained with $\Omega_{ori} + \Omega_{equ}$ outperforms one with $\Omega_{ori}$ by 1.1\% in terms of mAP under $IoU=[0.5:0.95]$. This clear improvement verifies the effectiveness of instance-switching strategy as data augmentation for object detection, and we owe the improvement to increase of diversity and keep of visual contextual coherence inherent in our instance-switching strategy.

To further verify effect of our instance-switching (IS) strategy, we make a statistical analysis. Specifically, we randomly pick $2K$ images with swappable instances (i.e., $\Omega_{t}^{2K}$ and $\Omega_{v}^{2K}$) from training ($\Omega_{t}$) and validation ($\Omega_{v}$) sets of MS COCO 2017, respectively. Then, we perform $1K$ times IS on $\Omega_{t}^{2K}$ and $\Omega_{v}^{2K}$. The image sets containing $2K$ switched instances of $\Omega_{t}^{2K}$ and $\Omega_{v}^{2K}$ are indicted by $\Omega_{t}^{IS}$ and $\Omega_{v}^{IS}$, respectively. As listed in Table~\ref{analysis}, Faster R-CNN trained on $\Omega_{t}$ misses 76 and 117 more instances on $\Omega_{t}^{IS}$ and $\Omega_{v}^{IS}$ than those on $\Omega_{t}^{2K}$ and $\Omega_{v}^{2K}$, respectively. In contrast, Faster R-CNN trained on $\Omega_{t} \bigcup \Omega_{t}^{IS}$ can reduce number of the missed instances, specifically on $\Omega_{t}^{IS}$ and $\Omega_{v}^{IS}$. The result indicates our IS can change the context of switched instances, thereby increasing the diversity of samples and leading the missing of pre-trained detector for switched instances, and it can be well suppressed by training detector on $\Omega_{t} \bigcup \Omega_{t}^{IS}$.

\begin{table*}[!t]
	\begin{center}
		\footnotesize
		\caption{Results (\%) of three state-of-the-art detectors (i.e., FPN~\cite{lin2017feature}, Mask R-CNN~\cite{he2017mask} and SNIPER~\cite{singh2018sniper}) under various augmentation methods. $\ast$ indicates no horizontal flipping, and $\times 2$ means two times training epochs, which is regarded as training-time augmentation~\cite{HeGDG18}.}\label{comparison}
		\begin{tabular}{l|c|ccc|ccc|ccc|ccc}
			\hline
			\multirow{2}{*}{Training sets} & \multirow{2}{*}{Detectors}  & \multicolumn{3}{c|}{ Avg.Precision, IOU:} & \multicolumn{3}{c|}{Avg.Precision, Area:} & \multicolumn{3}{c|}{Avg.Recall, \#Det:} & \multicolumn{3}{c}{Avg.Recall, Area:}\\
			\cline{3-14}
			& & 0.5:0.95 & 0.50 & 0.75 & Small & Med. & Large & 1 & 10 & 100 & Small & Med. &  Large\\
			\hline
			$\Omega_{ori}^{\ast}$ & \multirow{6}{*}{\tabincell{c}{FPN~\\\cite{lin2017feature}}} & 38.1 & 59.1 & 41.3 & 20.7 &  42.0 & 51.1 & 31.6 & 49.3 & 51.5 & 31.1 & 55.7 & 66.7\\
			$\Omega_{ori}$ & & 38.6 & 60.4 & 41.6 & 22.3 & 42.8 & 50.0 & 31.8 &  50.6 & 53.2 & 34.5 & 57.7 & 66.8\\
			$\Omega_{PSIS}^{\ast}$ &  & 38.7 & 59.7 & 41.8 & 21.6 & 43.0 & 51.7 & 32.0 &  50.0 & 52.3 & 32.3 & 56.4 & 67.6\\
			$\Omega_{PSIS}$ &  & 39.8 & 61.0 & 43.4 & 22.7 & 44.2 & 52.1 & 32.6 & 51.1 &  53.6 & 34.8 & 59.0 & 68.5\\
			$\Omega_{ori}\times 2$ &   & 39.4 & 60.7 & 43.0 & 22.1 & 43.6 & 52.1 & 32.5 & 51.0 &  53.4 & 33.6 &  57.6 & 68.6\\
			$\Omega_{PSIS}\times 2$ &   & \textbf{40.2} & \textbf{61.1} & \textbf{44.2} & \textbf{22.3} & \textbf{45.7} & \textbf{51.6} & \textbf{32.6} & \textbf{51.2} & \textbf{53.6} & \textbf{33.6} & \textbf{58.9} & \textbf{68.8}\\
			\hline
			$\Omega_{ori}$ & \multirow{4}{*}{\tabincell{c}{Mask R-CNN~\\\cite{he2017mask}}}  & 39.4 & 61.0 & 43.3 & 23.1 & 43.7 & 51.3 & 32.3 & 51.5 &  54.3 & 34.9 & 58.7 & 68.5\\
			$\Omega_{PSIS}$&  & 40.7 & 61.8 & 44.5 & 23.4 & 45.2 & 53.0 & 33.3 & 52.8 & 55.4 &  35.5 & 59.7 & 70.3\\
			$\Omega_{ori} \times 2$& & 40.4 & 61.6 & 44.2 & 22.3 & 44.8 & 52.9 & 33.1 & 52.0 & 54.5 & 34.7 & 58.8 & 69.5\\
			$\Omega_{PSIS} \times 2$& & \textbf{41.2} & \textbf{62.5} & \textbf{45.4} & \textbf{23.7} & \textbf{46.0} & \textbf{53.6} & \textbf{33.4} & \textbf{52.9} & \textbf{55.5} & \textbf{36.2}  & \textbf{60.0} & \textbf{70.3}\\
			\hline
			$\Omega_{ori}$&\multirow{2}{*}{\tabincell{c}{SNIPER ~\\\cite{singh2018sniper}}}  &  43.4 & 62.8 & 48.8 & 27.4 & 45.2 & 56.2 & N/A & N/A & N/A & N/A & N/A & N/A\\
			$\Omega_{PSIS}$&  & \textbf{44.2} & \textbf{63.5} & \textbf{49.3} & \textbf{29.3} & \textbf{46.2} & \textbf{57.1} & \textbf{35.0} &\textbf{60.1} & \textbf{65.9} &  \textbf{50.4} & \textbf{70.4} & \textbf{78.0}\\
			\hline
		\end{tabular}\vspace{-.5cm}
	\end{center}
\end{table*}

\vspace{-.3cm}
\subsubsection{Instance Balance Enhancement}~\label{sec4.1.2}\vspace{-.5cm}

\begin{table}[!t]
	\centering
	\scriptsize
	\caption{Statistical analysis of missed instances associated with $2K$ switched ones. Numbers in brackets indicate increased or decreased missed instances.}\label{analysis}\vspace{.2cm}
	\begin{tabular}{c|c|c|c|c}
		\hline
		&  $\Omega_{t}^{2K}$  & $\Omega_{t}^{IS}$ & $\Omega_{v}^{2K}$ & $\Omega_{v}^{IS}$\\
		\hline
	    Trained on $\Omega_{t}$ &   146  & 222 (+76) & 614 & 731 (+117) \\
		Trained on $\Omega_{t} \bigcup \Omega_{t}^{IS}$ &   133 (-13)  & 142 (-80) & 607 (-7) & 634 (-97) \\
		\hline
	\end{tabular}
\end{table}

\begin{figure}[!t]
	\begin{center}
		\includegraphics[width=0.8\linewidth]{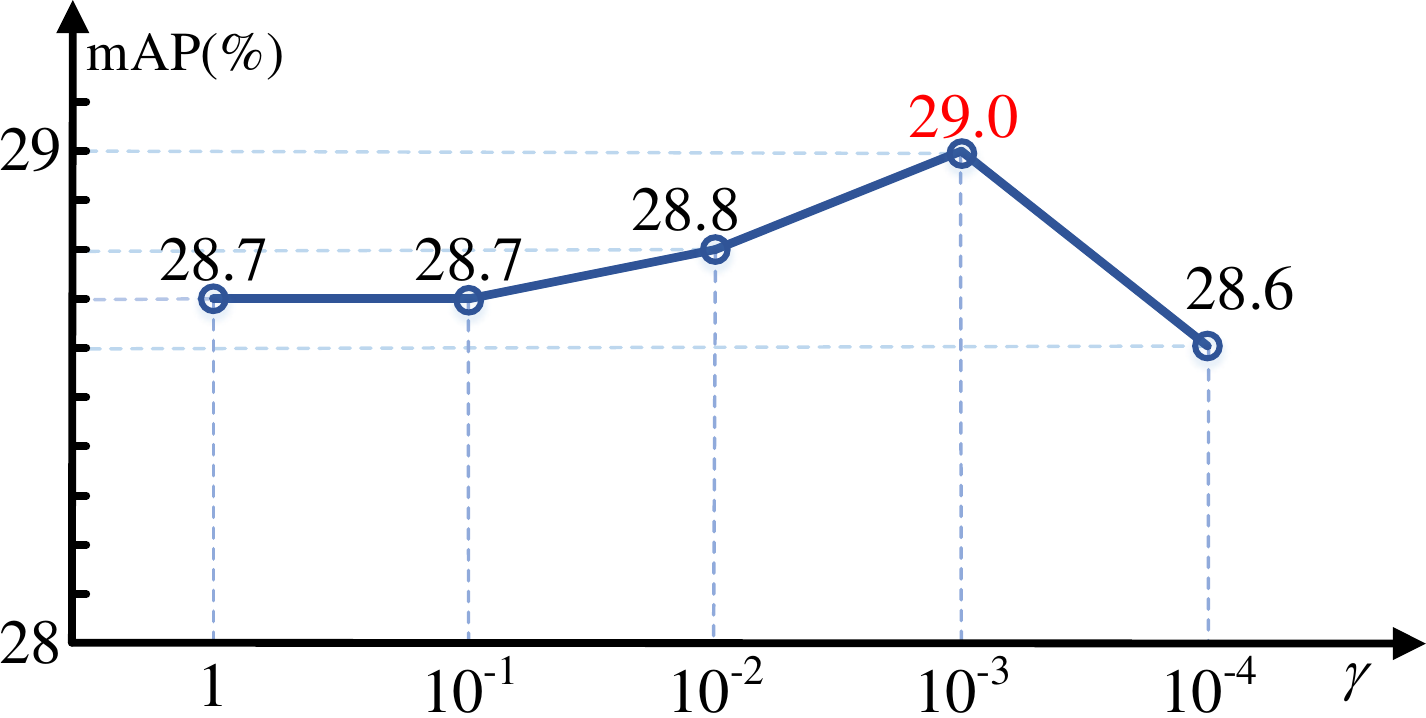}\\
		\caption{Results of loss $\ell_{CB}$ with various values of $\gamma$ by training Faster R-CNN on $\Omega_{ori} + \Omega_{uni}$. Note that $\gamma=1$ is the classical softmax loss. }\label{cbloss}
	\end{center}\vspace{-.5cm}
\end{figure}

Our PSIS introduces two strategies to enhance instance balance, including selective re-sampling strategy and class-balanced loss. As shown in Figure~\ref{rda} (b), our selective  re-sampling strategy is helpful to generate an approximately uniform training dataset $\Omega_{uni}$, which has the same size (i.e., 118k) with $\Omega_{ori}$ and $\Omega_{equ}$. As listed in Table~\ref{abstudy}, $\Omega_{uni}$ brings 1.4\% and 0.3\% gains over  $\Omega_{ori}$ and $\Omega_{equ}$ in terms of mAP under $IoU=[0.5:0.95]$, respectively. Combining with class-balanced loss $\ell_{CB}$, detection performance can further be improved to 29.0\%. Through introducing instance balance enhancement into our instance-switching strategy, we obtain 0.6\% improvement over $\Omega_{equ}$ that performs instance-switching in a non-uniform manner, indicating instance balance is an important property of dataset for better detection. Furthermore, we evaluate the effect of parameter $\gamma$ on class-balanced loss $\ell_{CB}$. In general, $\gamma \rightarrow 0$ indicates re-weighting instances in much larger inverse proportion to instance number. As illustrated in Figure~\ref{cbloss}, $\gamma=10^{-3}$ achieves the best result, outperforming $\gamma=1$ (i.e., the classical softmax loss) by 0.3\%. The smaller $\gamma$ ($\gamma < 10^{-3}$) leads performance decline. Therefore, we set $\gamma=10^{-3}$ in the following experiments.

\vspace{-.3cm}
\subsubsection{Progressive Instance-switching}~\label{sec4.1.3}\vspace{-.5cm}

\begin{figure}[!t]
	\begin{center}
		\includegraphics[width=0.8\linewidth]{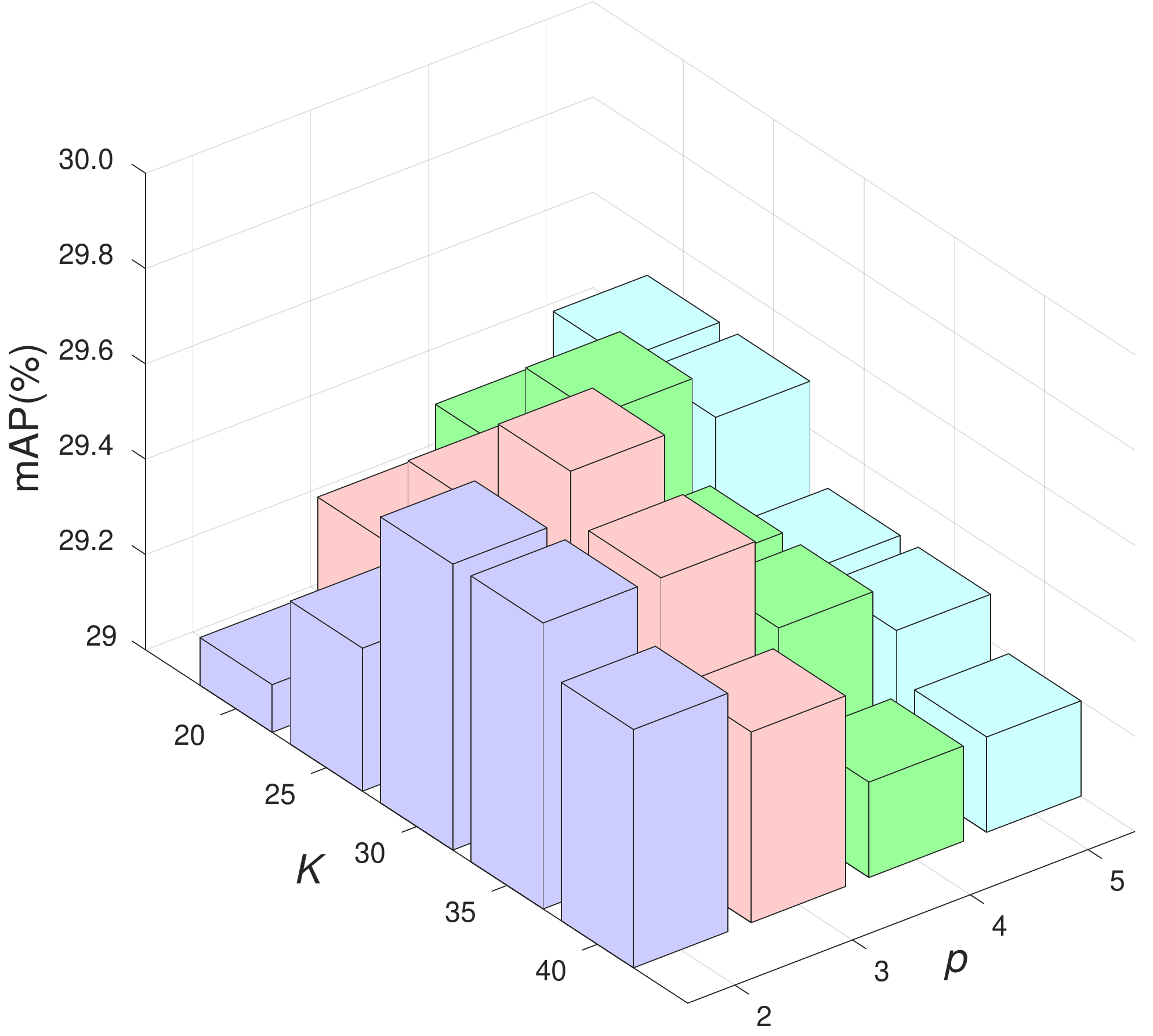}\\
		\caption{Results of $\Omega_{PSIS}$ using progressive instance-switching with various $K$ and $p$.}\label{PApk}
	\end{center}\vspace{-.5cm}
\end{figure}

To take the class importance into account, we propose a progressive instance-switching method for data augmentation. As listed in the bottom of Table~\ref{abstudy}, our progressive manner ($\Omega_{PSIS}$) brings 0.7\% gain in terms of mAP under $IoU=[0.5:0.95]$, demonstrating consideration of class importance is helpful to augment data for object detection. Note that our PSIS can achieve in total 2.4\% improvement over the original dataset $\Omega_{ori}$ under a strong detector (e.g., Faster R-CNN + ResNet-101). In addition, our progressive instance-switching (\ref{PIS}) involves two parameters, i.e., class number $K$ and image percentage $p$ to be augmented. Figure~\ref{PApk} shows the results of $\Omega_{PSIS}$ using progressive instance-switching with various $K$ and $p$. From it we can see that $\Omega_{PSIS}$  with $K=30,\,p=3$ obtains the best performance. Furthermore, generation of training images for more classes with larger proportions leads performance decline. It may be account for breaking of instance balance and over-fitting caused by over-sampling. Finally, our $\Omega_{PSIS}$ involves about 283k training images in total, including $\Omega_{ori}$ of 118k, $\Omega_{uni}$ of 118k and $\Omega_{aug}$ of 47k.

\vspace{.1cm}
\subsection{Apply $\Omega_{PSIS}$ to State-of-the-art Detectors}~\label{sec4.2}\vspace{-.1cm}

In above subsection, we have generated an augmented dataset $\Omega_{PSIS}$ based on Faster R-CNN, Then, we directly employ this dataset to train four state-of-the-art detectors (i.e.,  FPN~\cite{lin2017feature}, Mask R-CNN~\cite{he2017mask}, BlitzNet~\cite{dvornik2017blitznet} and SNIPER~\cite{singh2018sniper}), and report results on test server for comparing with other augmentation methods.

\vspace{-.3cm}
\subsubsection{PSIS for FPN}~\label{sec4.2.1}\vspace{-.5cm}

We firstly adopt our $\Omega_{PSIS}$ to FPN~\cite{lin2017feature} with RoI Alignment layer~\cite{he2017mask} under Faster R-CNN framework~\cite{ren2015faster}. We employ ResNet-101 as backbone model, and train FPN following the same settings
in~\cite{mmdetection2018}. Specifically, we train FPN on $\Omega_{ori}$ within 12 epochs, and set the initial learning rate to $10^{-2}$, which is decreased with a factor of 10 after 8 and 11 epochs. Due to $\Omega_{PSIS}$ containing more images, FPN is trained on $\Omega_{PSIS}$ with 18 epochs and learning rate is decreased after 13 and 16 epochs. Here we compare with two widely used data augmentation methods, i.e., \emph{horizontal flipping} and \emph{training-time augmentation}~\cite{HeGDG18}.

The comparison results are given in Table~\ref{comparison}, from it we can see that our PSIS and horizontal flipping bring $0.6\%$ and $0.5\%$ gains over non-augmented one, respectively. We further  combine our PSIS with horizontal flipping, which achieves 39.8\% mAP under $IoU=[0.5:0.95]$, outperforming horizontal flipping and non-augmented one by $1.2\%$ and $1.7\%$, respectively. We owe these gains to increase of sample diversity inherent in our PSIS. Besides, we also compare with training-time augmentation method (i.e., $2\times$ training epochs). Note that our PSIS without training-time augmentation is superior to training-time augmentation method by 0.4\%. By performing training-time augmentation, our PSIS further achieves 0.8\% improvement. Above results clearly demonstrate our PSIS is superior and complementary to horizontal flipping and training-time augmentation methods.

\vspace{-.3cm}
\subsubsection{PSIS for Mask R-CNN}~\label{sec4.2.2}\vspace{-.5cm}

We evaluate our PSIS using Mask R-CNN~\cite{he2017mask}, which also exploits mask information to improve object detection accuracy while achieving state-of-the-art performance. Here, we employ ResNet-101 as backbone model, and train Mask R-CNN using~\cite{mmdetection2018} and employing the same settings with FPN. We compare our PSIS with two augmentation manners, i.e., \emph{segmentation mask} and \emph{training-time augmentation}. The results are compared in Table~\ref{comparison}, where we can see that  Mask R-CNN trained with our $\Omega_{PSIS}$ consistently outperforms one with the original dataset $\Omega_{ori}$ under all evaluation metrics, although Mask R-CNN exploits instance masks as well. In particular, our PSIS brings 1.3\% gains under $IoU=[0.5:0.95]$, demonstrating the complementarity between mask branch of Mask R-CNN and our PSIS. For training-time augmentation, $\Omega_{PSIS}\times 1$ is better than $\Omega_{ori}$ with 2 times training epochs. Combining with training-time augmentation, PSIS is further improved by 0.5\% . Above results show again our PSIS is superior and complementary to training-time augmentation method.

\vspace{-.3cm}
\subsubsection{PSIS for SNIPER}~\label{sec4.2.3}\vspace{-.5cm}

SNIPER~\cite{singh2018sniper} is a recently proposed high-performance detector, which handles object detection in an efficient multi-scale manner. To verify the effectiveness of our PSIS under \emph{multi-scale training strategy}, we employ ResNet-101 as backbone model and train SNIPER on $\Omega_{ori}$ and $\Omega_{PSIS}$ following the experimental settings in~\cite{singh2018sniper} except batch-size and training epochs. Due to the limited computing resources, we use half of batch-size and double training epochs in the original settings, and we do not use negative chip mining for~\cite{singh2018sniper}. As compared in Table~\ref{comparison}, SNIPER trained on our $\Omega_{PSIS}$ achieves 0.8\% gain over one with the original training dataset, showing the proposed PSIS is complementary to multi-scale strategy and it can be flexibly adopted to multi-scale training/testing detectors.

\begin{table*}[!t]
	\begin{center}
		\footnotesize
		\caption{Results (\%) of FPN and Mask R-CNN using different training sets and epochs on MS COCO. $\Omega_{ori}^{\dagger}$ and $\Omega_{ori}^{\ddagger}$: training on $\Omega_{ori}$ within 18 and 44  epochs.}\label{fair}
		\begin{tabular}{l|c|ccc|ccc|ccc|ccc}
			\hline
			\multirow{2}{*}{Training sets} & \multirow{2}{*}{Detectors}  & \multicolumn{3}{c|}{Avg.Precision, IOU:} & \multicolumn{3}{c|}{Avg.Precision, Area:} & \multicolumn{3}{c|}{Avg.Recall, \#Det:} & \multicolumn{3}{c}{Avg.Recall, Area:}\\
			\cline{3-14}
			& & 0.5:0.95 & 0.50 & 0.75 & Small & Med. & Large & 1 & 10 & 100 & Small & Med. & Large\\
			\hline
			$\Omega_{ori}^{\dagger}$  &\multirow{3}{*}{\tabincell{c}{FPN~\\\cite{lin2017feature}}} & 39.0  & 60.2 & 42.6 & 22.5 & 43.2 & 50.7 & 32.4 & \textbf{51.5} & 54.1 & 34.8 & \tiny58.4 & \textbf{68.6}
\\
			$\Omega_{ori}^{\ddagger}$ & & 38.9  & 60.3 & 42.6 & 22.6 & 43.3 & 50.0 & 32.3 & \textbf{51.5} & \textbf{54.2} & \textbf{35.1} & 58.4 & 68.3\\
            $\Omega_{PSIS}$ &  & \textbf{39.8} & \textbf{61.0} & \textbf{43.4} & \textbf{22.7} & \textbf{44.2} & \textbf{52.1} & \textbf{32.6} & 51.1 & 53.6 & 34.8 & \textbf{59.0} & 68.5\\
			\hline
			$\Omega_{ori}^{\dagger}$ & \multirow{3}{*}{\tabincell{c}{Mask R-CNN~\\\cite{he2017mask}}} & 40.2  & 61.2 & 44.4 & 23.2 & 44.1 & 51.8 & 33.0 & 52.3 & 54.9 & \textbf{35.5} & 59.2 & 69.2\\
			$\Omega_{ori}^{\ddagger}$&  & 40.3  & 61.3 & 44.3 & \textbf{23.7} & 44.3 & 51.9 & 33.0 & 52.3 & 54.9 & 35.4 & 59.1 & 69.2\\
            $\Omega_{PSIS}$&  & \textbf{40.7} & \textbf{61.8} & \textbf{44.5} & 23.4 & \textbf{45.2} & \textbf{53.0} & \textbf{33.3} & \textbf{52.8} & \textbf{55.4} & \textbf{35.5} & \textbf{59.7} & \textbf{70.3}\\
			\hline
		\end{tabular}\vspace{-.3cm}
	\end{center}
\end{table*}

\begin{table*}[!t]
	\begin{center}
		\footnotesize
		\caption{Results (\%) of Mask R-CNN training with different datasets on instance segmentation task of MS COCO 2017.}\label{IScomparison}
		\begin{tabular}{l|c|ccc|ccc|ccc|ccc}
			\hline
			\multirow{2}{*}{Training sets} & \multirow{2}{*}{Method}  & \multicolumn{3}{c|}{Avg.Precision, IOU:} & \multicolumn{3}{c|}{Avg.Precision, Area:} & \multicolumn{3}{c|}{Avg.Recall, \#Det:} & \multicolumn{3}{c}{Avg.Recall, Area:}\\
			\cline{3-14}
			& &  0.5:0.95 & 0.50 & 0.75 & Small & Med. & Large & 1 & 10 & 100 & Small & Med. & Large\\
			\hline
			$\Omega_{ori}$ & \multirow{4}{*}{\tabincell{c}{Mask R-CNN~\\\cite{he2017mask}}} & 35.9 & 57.7 & 38.4 & 19.2 & 39.7 & 49.7 & 30.5 & 47.3 & 49.6 & 29.7 & 53.8 & 65.8\\
			$\Omega_{PSIS}$ &  & 36.7 & 58.4 & 39.4 & 19.0 & 40.6 & 50.2 & 31.0 & 48.2 & 50.3 &  29.8 & 54.4 & 66.9\\
			$\Omega_{ori}\times 2$ &  & 36.6 & 58.2 & 39.2 & 18.5 & 40.3 & 50.4 & 31.0 & 47.7 & 49.7 & 29.5 & 53.5 & 66.6\\
			$\Omega_{PSIS}\times2$  & & \textbf{37.1} &  \textbf{58.8} & \textbf{39.9} & \textbf{19.3} & \textbf{41.2} & \textbf{50.8} & \textbf{31.1} & \textbf{47.7} & \textbf{50.4} & \textbf{30.2} & \textbf{54.5} & \textbf{67.0}\\
			\hline
		\end{tabular}\vspace{-.3cm}
	\end{center}
\end{table*}

\begin{table*}[!t]
	\begin{center}
		\footnotesize
		\caption{Results (\%) of pre-trained Faster R-CNN  using MS COCO on PASCAL VOC 2007 test set. $\Omega_{ori}^{coco-}$ and $\Omega_{PSIS}^{coco-}$ outperforms $\Omega_{ori}^{coco}$ and $\Omega_{PSIS}^{coco}$, respectively, as latter involve many irrelevant classes.}\label{voc1}\vspace{.1cm}
		\begin{tabular}{l|cccccccccccccccccccc|c}
			\hline
			{\tiny{Training sets}} & \tiny{aero}&\tiny{bike}&\tiny{bird}&\tiny{boat}&\tiny{bottle}
            &\tiny{bus}&\tiny{car}&\tiny{cat}&\tiny{chair}&\tiny{cow}&\tiny{table}&\tiny{dog}&\tiny{horse}
            &\tiny{mbike}&\tiny{person}&\tiny{plant}&\tiny{sheep}&\tiny{sofa}&\tiny{train}&\tiny{tv}&\tiny{mAP}\\
            \hline
			$\Omega_{ori}^{coco}$ & \tiny81.4 & \tiny80.6 & \tiny81.1 & \tiny66.0 & \tiny73.8 & \tiny85.1 & \tiny81.6 & \tiny87.1 & \tiny64.4 & \tiny86.0 & \tiny59.9 & \tiny85.6 & \tiny87.7 & \tiny85.3 & \tiny85.1 & \tiny49.6 & \tiny84.2 & \tiny73.7 & \tiny87.8 & \tiny80.5 &\tiny78.3\\
			$\Omega_{PSIS}^{coco}$ & \tiny\textbf{82.2} & \tiny\textbf{82.4} & \tiny\textbf{83.9} & \tiny\textbf{67.8} & \tiny\textbf{74.3} & \tiny\textbf{85.8} & \tiny\textbf{82.2} & \tiny\textbf{87.9} & \tiny\textbf{65.1} & \tiny\textbf{87.5} & \tiny\textbf{61.8} & \tiny\textbf{85.7} & \tiny\textbf{89.3} & \tiny\textbf{86.8} & \tiny\textbf{86.1} & \tiny\textbf{52.5} & \tiny\textbf{85.0} & \tiny\textbf{75.3} & \tiny\textbf{88.5} & \tiny\textbf{81.2} & \tiny\textbf{79.6}\\
			\hline
			$\Omega_{ori}^{coco-}$ & \tiny82.1 & \tiny82.0 & \tiny83.3 & \tiny67.4 & \tiny74.3 & \tiny87.2 & \tiny82.9 & \tiny88.0 & \tiny64.1 & \tiny87.8 & \tiny62.1 & \tiny86.5 & \tiny89.1 & \tiny87.5 & \tiny85.2 & \tiny52.4 & \tiny85.0 & \tiny76.2 & \tiny87.3 & \tiny80.1 &\tiny79.5\\
			$\Omega_{PSIS}^{coco-}$ & \tiny\textbf{83.9} & \tiny\textbf{84.0} & \tiny\textbf{85.0} & \tiny\textbf{69.3} & \tiny\textbf{76.2} & \tiny\textbf{88.9} & \tiny\textbf{84.0} & \tiny\textbf{89.2} & \tiny\textbf{67.2} & \tiny\textbf{88.9} & \tiny\textbf{64.0} & \tiny\textbf{86.6} & \tiny\textbf{90.6} & \tiny\textbf{88.3} & \tiny\textbf{87.4} & \tiny\textbf{54.0} & \tiny\textbf{86.1} & \tiny\textbf{76.9} & \tiny\textbf{89.6} & \tiny\textbf{82.3} & \tiny\textbf{81.1}\\
            \hline
		\end{tabular}\vspace{-.3cm}
	\end{center}
\end{table*}

\vspace{-.3cm}
\subsubsection{PSIS for BlitzNet}~\label{sec4.2.4}\vspace{-.5cm}

Recently, Dvornik \etal propose a context-based data augmentation (\emph{Context-DA}) method for object detection~\cite{dvornik2018importance}, where a real-time single-shot detector (dubbed by BlitzNet~\cite{dvornik2017blitznet}) is used to evaluate performance of Context-DA. In comparison to Context-DA~\cite{dvornik2018importance}, we also adopt our PSIS to BlitzNet. Following the settings in ~\cite{dvornik2018importance}, we train BlitzNet using  ResNet-50 as backbone model, while set batch-size and training epochs to half of and 2 times original ones. Note that the original BlitzNet runs on MS COCO 2014, so we rebuild our  $\Omega_{PSIS}$ based on the training set of MS COCO 2014. The results are compared in Table~\ref{BlitzNet0}. Clearly, both Context-DA and our PSIS improve the original dataset $\Omega_{ori}$. Meanwhile, our PSIS outperforms Context-DA by 2.8\% under $IoU=[0.5:0.95]$. We owe this improvement to that our PSIS can preserve contextual coherence in the original images and benefit from more appropriate visual context.

\begin{table}[!t]
	\begin{center}
		\footnotesize
		\caption{Results (\%) of BlitzNet~\cite{dvornik2017blitznet} training with different datasets on MS COCO 2014.}\label{BlitzNet0}	
		\begin{tabular}{l|ccc|ccc}
			\hline
			\multirow{2}{*}{Training set}  & \multicolumn{3}{c|}{Avg.Precision, IOU:} & \multicolumn{3}{c}{Avg.Precision, Area:} \\
			\cline{2-7}
			& 0.5:0.95 & 0.50 & 0.75 &  Small &  Med. & Large \\
			\hline
			$\Omega_{ori}$  & 27.3 & 46.0 & 28.1 & 10.7 & 26.8 & 46.0\\
			Context-DA~\cite{dvornik2018importance}  & 28.0 & 46.7 & 28.9 & 10.7 & 27.8 &  47.0\\
			$\Omega_{PSIS}$  & \textbf{30.8} & \textbf{50.0} & \textbf{32.2} & \textbf{12.6} & \textbf{31.0} & \textbf{50.2} \\
			\hline
		\end{tabular}
	\end{center}\vspace{-.5cm}
\end{table}

\vspace{.15cm}
\subsection{Longer Training Time on $\Omega_{ori}$}\vspace{.15cm}

As described in Section ~\ref{sec4.2}, our PSIS exploits more training epochs. To evaluate its effect, we also train FPN~\cite{lin2017feature} and Mask R-CNN~\cite{he2017mask} on the original set $\Omega_{ori}$ longer time. Specifically, we train the detectors within 18 (same with $\Omega_{PSIS}$) and 44 (considering effects of both more data and longer training) epochs on $\Omega_{ori}$, which are indicted by $\Omega_{ori}^{\dagger}$ and $\Omega_{ori}^{\ddagger}$, respectively. The results are given in Table~\ref{fair}, where our PSIS ($\Omega_{PSIS}$) outperforms both $\Omega_{ori}^{\dagger}$ and $\Omega_{ori}^{\ddagger}$. Moreover, $\Omega_{ori}^{\ddagger}$ brings little gains over  $\Omega_{ori}^{\dagger}$, indicating more training epochs lead over-fitting on $\Omega_{ori}$. Our PSIS avoids over-fitting by increasing diversity of samples and benefits from longer training-time.

\begin{table*}[!t]
	\begin{center}
	    \footnotesize
		\caption{Results (\%) of different methods on PASCAL VOC 2007 test set. $\Omega_{ori}^{F}$ and $\Omega_{PSIS}^{F}$ indicate training Faster R-CNN~\cite{ren2015faster} on $\Omega_{ori}$ and $\Omega_{PSIS}$, respectively. $\Omega_{ori}^{B}$, $\Omega_{DA}^{B}$ and $\Omega_{PSIS}^{B}$ indicate training BlitzNet~\cite{dvornik2017blitznet} on $\Omega_{ori}$, $\Omega_{DA}$  and $\Omega_{PSIS}$, respectively.}\label{voc2}\vspace{.1cm}
		\begin{tabular}{l|cccccccccccccccccccc|c}
			\hline
			{\tiny{Training sets}} &\tiny{aero}&\tiny{bike}&\tiny{bird}&\tiny{boat}&\tiny{bottle}
            &\tiny{bus}&\tiny{car}&\tiny{cat}&\tiny{chair}&\tiny{cow}&\tiny{table}&\tiny{dog}&\tiny{horse}
            &\tiny{mbike}&\tiny{person}&\tiny{plant}&\tiny{sheep}&\tiny{sofa}&\tiny{train}&\tiny{tv}&\tiny{mAP}\\
			\hline
			$\Omega_{ori}^{F}$ & \tiny60.9 & \tiny75.8 & \tiny65.8 & \tiny48.4 & \tiny57.0 & \tiny74.1 & \tiny70.2 & \tiny83.0 & \tiny40.6 & \tiny74.7 & \tiny56.3 & \tiny80.1 & \tiny77.7 & \tiny69.8 & \tiny67.6 & \tiny35.0 & \tiny66.9 & \tiny58.0 & \tiny65.5 & \tiny57.2 &\tiny64.2\\
			$\Omega_{PSIS}^{F}$ & \tiny\textbf{63.2} & \tiny\textbf{76.4} & \tiny\textbf{67.0} & \tiny\textbf{50.6} & \tiny\textbf{57.8} & \tiny\textbf{74.3} & \tiny\textbf{71.0} & \tiny\textbf{83.3} & \tiny\textbf{43.3} & \tiny\textbf{76.5} & \tiny\textbf{56.8} & \tiny\textbf{82.5} & \tiny\textbf{78.0} & \tiny\textbf{69.9} & \tiny\textbf{68.3} & \tiny\textbf{35.5} & \tiny\textbf{68.8} & \tiny\textbf{58.2} & \tiny\textbf{67.2} & \tiny\textbf{59.9} &\tiny\textbf{65.4}\\
			\hline
			${\Omega_{ori}^{B}}$ & \tiny63.6 & \tiny73.3 & \tiny63.2 & \tiny57.0 & \tiny31.5 & \tiny76.0 & \tiny71.5 & \tiny79.9 & \tiny40.0 & \tiny71.6 & \tiny61.4 & \tiny74.6 & \tiny80.9 & \tiny70.4 & \tiny67.9 & \tiny36.5 & \tiny64.9 & \tiny63.0 & \tiny79.3 & \tiny64.7 &\tiny64.6 \\
			${\Omega_{DA}^{B}}$& \tiny66.8 & \tiny75.3 & \tiny65.9 & \tiny57.2 & \tiny33.1 & \tiny75.0 & \tiny72.4 & \tiny79.6 & \tiny40.6 & \tiny73.9 & \tiny63.7 & \tiny\textbf{77.1} & \tiny\textbf{81.4} & \tiny\textbf{71.8} & \tiny68.1 & \tiny37.9 & \tiny67.6 & \tiny64.7 & \tiny81.2 & \tiny65.5 &\tiny65.9\\
           ${\Omega_{PSIS}^{B}}$& \tiny\textbf{67.6} & \tiny\textbf{76.4} & \tiny\textbf{66.3} & \tiny\textbf{57.7} & \tiny\textbf{33.5} & \tiny\textbf{75.5} & \tiny\textbf{73.9} & \tiny\textbf{80.8} & \tiny\textbf{42.6} & \tiny\textbf{74.5} & \tiny\textbf{65.5} & \tiny77.0 & \tiny\textbf{81.4} & \tiny71.6 & \tiny\textbf{69.3} & \tiny\textbf{40.2} & \tiny\textbf{68.4} & \tiny\textbf{65.1} & \tiny\textbf{81.7} & \tiny\textbf{66.3} &\tiny\textbf{66.8}\\
			\hline
		\end{tabular}\vspace{-.3cm}
	\end{center}
\end{table*}

Due to limited computing resources, we use \emph{half of batch-size} and \emph{double training epochs} of original settings when training SNIPER and BlitzNet on $\Omega_{PSIS}$. To assess its effect, we report the results of SNIPER and BlitzNet trained on $\Omega_{ori}$ with the same settings in Table~\ref{sniper} and Table~\ref{BlitzNet1}. They obtain 43.2\% and 27.2\% at IoU=[0.5:0.95], respectively. For SNIPER, our PSIS (44.2\%) can achieve 1\% gain. Meanwhile, our PSIS (30.8\%) can achieve 3.6\% gains for BlitzNet.

\vspace{.1cm}
\subsection{Generalization to Instance Segmentation}~\label{sec4.3}\vspace{-.2cm}

We verify the generalization ability of our PSIS on instance segmentation task of MS COCO 2017. To this end, we train Mask R-CNN~\cite{he2017mask} (ResNet-101) following the exactly same settings in Section~\ref{sec4.2.2}. As shown in Table~\ref{IScomparison}, our PSIS can improve the original training dataset over $0.8\%$ and $0.5\%$ under $IoU=[0.5:0.95]$ within  $\times 1$ and $\times 2$ training epochs, respectively. PSIS further achieves $0.4\%$ gain in training-time augmentation manner. As for Mask R-CNN, it exploits instance masks to simultaneously perform detection and segmentation through a multi-task loss for obtaining better performance. Different from it, our PSIS employs instance masks to augment training data, and the results clearly show our PSIS offers a new and complementary way to use instance masks for improving both detection and segmentation performance. Besides, above results indicate our PSIS is independent on pre-defined detector, and can generalize well to various detectors and tasks.

\vspace{.15cm}
\subsection{Generalization to Small-scale PASCAL VOC}\vspace{.15cm}

We further verify the generalization ability of PSIS by adopting Faster R-CNN~\cite{ren2015faster} trained with MS COCO to test set of PASCAL VOC 2007~\cite{everingham2010pascal} without any fine-tuning. Here, we train Faster R-CNN on full MS COCO ($\Omega_{ori}^{coco}$/$\Omega_{PSIS}^{coco}$) or a subset of MS COCO ($\Omega_{ori}^{coco-}$/$\Omega_{PSIS}^{coco-}$). The latter only contains 20 classes those share with ones of PASCAL VOC. As shown in Table~\ref{voc1}, our PSIS respectively obtains 1.3\% and 1.6\% gains in term of mAP, comparing with ones trained with the original datasets $\Omega_{ori}^{coco}$ and $\Omega_{ori}^{coco-}$. Above results demonstrate the improvement achieved by PSIS on MS COCO can be generalized to other dataset.

Furthermore, we directly adopt PSIS to PASCAL VOC~\cite{everingham2010pascal}. Following the same experimental settings in Context-DA~\cite{dvornik2018importance}, we use PASCAL VOC 2012 training set that equips with segmentation annotations (including 1464 images) for training and test set of PASCAL VOC 2007 for testing. Meanwhile, Faster R-CNN~\cite{ren2015faster} and BlitzNet~\cite{dvornik2017blitznet} are used for evaluation. The results are compared in Table~\ref{voc2}. For Faster R-CNN, our PSIS obtains 1.2\% gains over the original training set $\Omega_{ori}^{F}$. For BlitzNet, PSIS outperforms $\Omega_{ori}^{B}$ and Context-DA~\cite{dvornik2018importance} ($\Omega_{DA}^{B}$) by 2.2\% and 0.9\%, respectively. These improvements clearly show our PSIS has good generalization ability to different datasets.

\section{Conclusion}

In this paper, we proposed a simple yet effective data augmentation for object detection, whose core is a progressive and selective instance-switching (PSIS) method for synthetic image generation. The proposed PSIS as data augmentation for object detection benefits several merits, i.e., increase of diversity of samples, keep of contextual coherence in the original images, no requirement of external datasets, and consideration of instance balance and class importance. Experimental results demonstrate the effectiveness of our PSIS against the existing data augmentation, including horizontal flipping and training time augmentation for FPN, segmentation masks and training time augmentation for Mask R-CNN, multi-scale training strategy for SNIPER, and Context-DA for BlitzNet. The improvement on both object detection and instance segmentation tasks suggest our proposed PSIS has the potential to improve the performance of other applications (i.e., keypoint detection), which will be investigated in future work.

{\small
\bibliographystyle{ieee}
\bibliography{egbib}
}

\clearpage

\section{Supplementary Material}

\begin{table*}[!t]
	\begin{center}
	 \footnotesize
	 \caption{Results (\%) of Faster R-CNN on validation set of MS COCO 2017 under different depth networks (e.g., ResNet-50, ResNet-101 and ResNet-152).}\label{deepnet}
		\begin{tabular}{l|c|ccc|ccc|ccc|ccc}
			\hline
			\multirow{2}{*}{ Training sets} & \multirow{2}{*}{Detector}  & \multicolumn{3}{c|}{ Avg.Precision, IOU:} & \multicolumn{3}{c|}{Avg.Precision, Area:} & \multicolumn{3}{c|}{ Avg.Recall, \#Det:} & \multicolumn{3}{c}{Avg.Recall, Area:}\\
			\cline{3-14}
			& & 0.5:0.95 & 0.50 & 0.75 &  Small & Med. & Large & 1 &   10 &  100 &  Small &  Med. &  Large\\
			\hline
			 $\Omega_{ori}$  & \multirow{2}{*}{\tabincell{c}{Faster R-CNN\\(ResNet-50)\cite{ren2015faster}}}  & 26.2 & 47.4 & 26.8 & 8.3 & 29.7 & 41.8 & 25.5 & 37.9 & 39.0 & 16.0 & 45.0 & 59.5\\
			 $\Omega_{PSIS}$ &  & \textbf{27.7} & \textbf{49.1} & \textbf{28.7} & \textbf{9.2} & \textbf{31.5} &  \textbf{44.2} & \textbf{26.5} & \textbf{39.3} &  \textbf{40.5} & \textbf{17.7} & \textbf{46.7} & \textbf{61.6}\\
			\hline
			 $\Omega_{ori}$  & \multirow{2}{*}{\tabincell{c}{Faster R-CNN\\(ResNet-101)\cite{ren2015faster}}}  & 27.3 & 48.6 & 27.5 & 8.8 & 30.4 & 43.1 & 25.8 & 38.5 & 39.3 & 16.3 & 44.5& 60.2\\
			 $\Omega_{PSIS}$ &  & \textbf{29.7} & \textbf{50.6} & \textbf{30.5} & \textbf{10.3} &  \textbf{33.4} & \textbf{48.0} & \textbf{27.6} &  \textbf{40.5} & \textbf{41.6} & \textbf{18.6} & \textbf{47.6} & \textbf{64.6} \\
			\hline
			 $\Omega_{ori}$  & \multirow{2}{*}{\tabincell{c}{Faster R-CNN\\(ResNet-152)\cite{ren2015faster}}}  &  28.0 & 49.0 & 28.8 & 9.2 & 31.3 & 44.9 & 26.6 & 39.7 & 40.7 & 18.0 & 46.9& 62.9 \\
			 $\Omega_{PSIS}$ &  & \textbf{30.3} & \textbf{51.7} & \textbf{31.4} & \textbf{10.5} & \textbf{33.8} &  \textbf{48.5} & \textbf{27.8} & \textbf{41.0} &  \textbf{42.2} & \textbf{18.9} & \textbf{48.5} & \textbf{64.8}\\
			 \hline
		\end{tabular}
	\end{center}
\end{table*}

\begin{figure*}[!t]
	\begin{center}
		\includegraphics[width=0.9\linewidth]{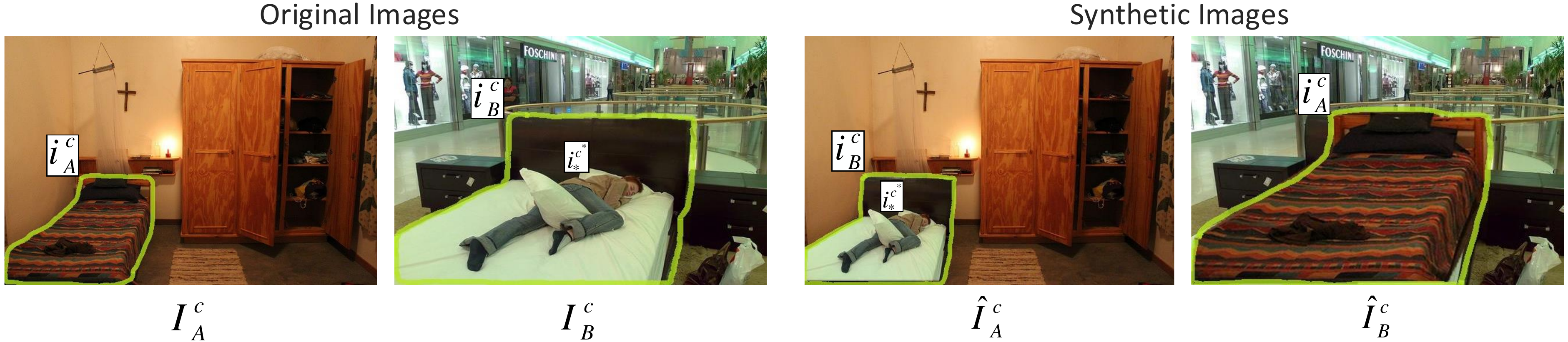}\\
		\caption{Switching and annotation for attaching instances.}\label{annupdate}
	\end{center}\vspace{-0.4cm}
\end{figure*}

In this supplementary file, we provide more qualitative and quantitative analysis for our proposed PSIS. Specifically, we further verify the effect of progressive instance-switching, and provide more examples of synthetic images generated by our IS strategy. Moreover, we evaluate the effect of backbone models with various depths. Finally, we assess the effects of training settings (i.e., batch-size, epochs and negative chip mining) on BlitzNet~\cite{dvornik2017blitznet} and SNIPER~\cite{singh2018sniper}.

\subsection{Illustration of Examples of Synthetic Images}~\label{imgshow}
\vspace{-.7cm}

\paragraph{Switching and Annotation for Attaching Instances.} Given a quadruple  $\{I_{A}^{c}$,$I_{B}^{c},i_{A}^{c},i_{B}^{c}\}$ in the candidate set $\Omega^{c}_{cad}$, our IS strategy switches the instances $i_{A}^{c},i_{B}^{c}$ to obtain a new synthetic image pair $\{\hat{I}_{A}^{c}$,$\hat{I}_{B}^{c}\}$. However, as shown in Figure~\ref{annupdate}, some instances (e.g., person $i^{c^\ast}_\ast$) may be attached on a switched instance (e.g., bed $i_{B}^{c}$). To preserve contextual coherence, our IS also switches and annotates the attaching instances during generation of the synthetic images.

\begin{figure*}[!t]
	\begin{center}
		\includegraphics[width=0.85\linewidth]{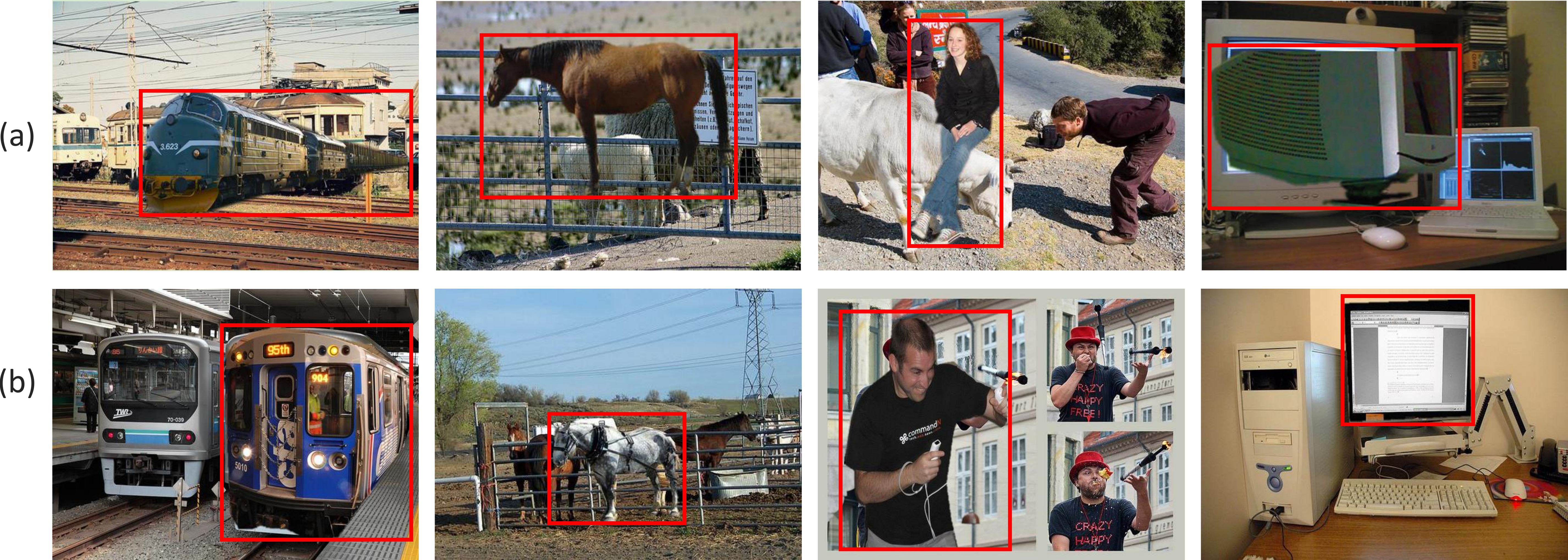}\\
		\caption{Comparison of synthetic images generated by (a) Context-DA and (b) our instance-switching strategy. The new instances are denoted in red boxes. Compared with Context-DA, our instance-switching strategy can better preserve contextual coherence in the original images and has better visual authenticity.}\label{cmp}
	\end{center}
\end{figure*}

\paragraph{Comparison with Context-DA~\cite{dvornik2018importance}.} We also compare the synthetic images generated by Context-DA and our instance-switching strategy. The synthetic images generated by Context-DA are copied from the original paper~\cite{dvornik2018importance}, and we pick the synthetic images generated by our instance-switching strategy, sharing the similar scenes with ones of Context-DA. The image examples are illustrated in Figure~\ref{cmp}, where we can see that our instance-switching strategy can better preserve contextual coherence in the original images in comparison to Context-DA. Meanwhile,  the synthetic images generated by our instance-switching strategy have better visual authenticity.

\begin{figure*}[!t]
	\begin{center}
		\includegraphics[width=0.65\linewidth]{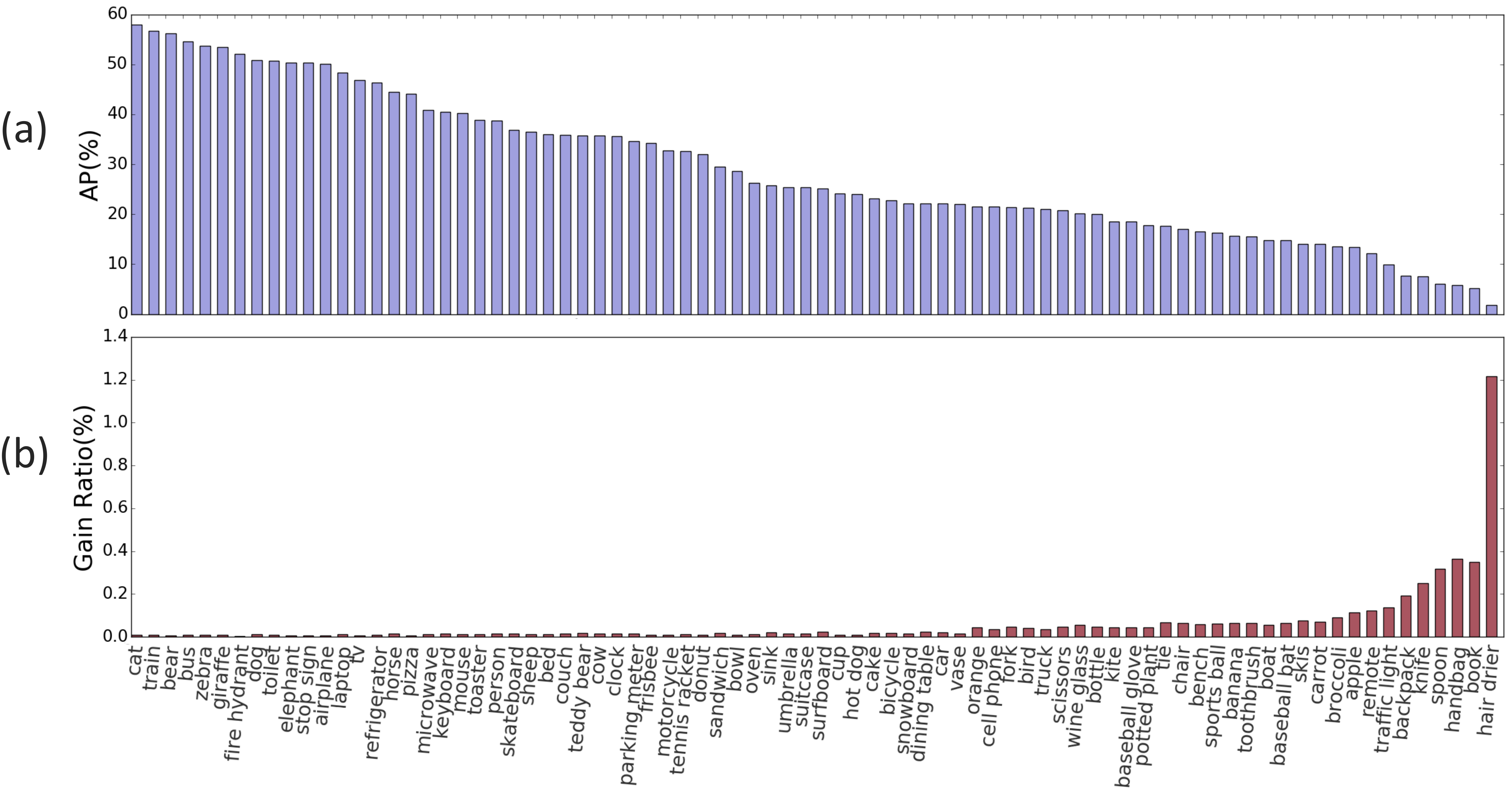}\\
		\caption{Results of Faster R-CNN (ResNet-101) on MS COCO dataset. (a) shows AP of each class before progressive instance-switching in descending order; (b) gives
			gain ratios of each class brought by progressive instance-switching.}\label{pis}
	\end{center}\vspace{-.7cm}
\end{figure*}

\paragraph{Examples of Synthetic Images Generated by Our IS}
Here we show some examples of synthetic images generated by our IS strategy. As illustrated in
Figure~\ref{good1}, the new (switched) instances are denoted in red boxes, and our instance-switching strategy can clearly preserve contextual coherence in the original images.

\begin{table*}[!t]
	\begin{center}
	 \footnotesize
	 \caption{Results (\%) of SNIPER (ResNet-101) with different training configurations on test-dev set of MS COCO 2017. The results of $\Omega_{ori}$ are copied from \cite{singh2018sniper}.}\label{sniper} 
		\begin{tabular}{l|c|ccc|ccc|ccc|ccc}
			\hline
			\multirow{2}{*}{ Training sets} & \multirow{2}{*}{Detector}  & \multicolumn{3}{c|}{ Avg.Precision, IOU:} & \multicolumn{3}{c|}{Avg.Precision, Area:} & \multicolumn{3}{c|}{ Avg.Recall, \#Det:} & \multicolumn{3}{c}{Avg.Recall, Area:}\\
			\cline{3-14}
			& & 0.5:0.95 & 0.50 & 0.75 &  Small & Med. & Large & 1 &   10 &  100 &  Small &  Med. &  Large\\
			\hline
			 $\Omega_{ori}^{\dagger}$  & \multirow{2}{*}{\tabincell{c}{SNIPER\\(ResNet-101)\cite{singh2018sniper}\\W/o neg}}  & 43.2 & 62.5 & 48.7 & 28.3 & 45.6 & 56.7 & 34.7 & 59.4 & 65.3 & 49.5 & 70.0 & 77.1\\
             $\Omega_{ori}$ &  &  43.4 & 62.8 & 48.8 & 27.4 & 45.2 & 56.2 & N/A & N/A & N/A & N/A & N/A & N/A\\
			 $\Omega_{PSIS}$&  & \textbf{44.2} & \textbf{63.5} & \textbf{49.3} & \textbf{29.3} & \textbf{46.2} & \textbf{57.1} & \textbf{35.0} &\textbf{60.1} & \textbf{65.9} &  \textbf{50.4} & \textbf{70.4} & \textbf{78.0}\\
            \hline
             $\Omega_{ori}^{\dagger}$  & \multirow{2}{*}{\tabincell{c}{SNIPER\\(ResNet-101)\cite{singh2018sniper}\\with neg}}  & 46.4 & 67.2 & 52.3 & 30.7 & 49.5 & 58.8 & 36.9 & 62.1 & 68.2 & 52.1 & 73.3 & 82.0\\
             $\Omega_{ori}$ &  &  46.5 & 67.5 & 52.2 & 30.0 & 49.4 & 58.4 & N/A & N/A & N/A & N/A & N/A & N/A\\
			 $\Omega_{PSIS}$&  & \textbf{47.1} & \textbf{68.5} & \textbf{52.8} & \textbf{32.1} & \textbf{50.2} & \textbf{59.0} & \textbf{37.4} &\textbf{63.2} & \textbf{69.0} &  \textbf{53.1} & \textbf{73.9} & \textbf{82.6}\\
			 \hline
		\end{tabular}
	\end{center}
\end{table*}

\begin{figure*}[!t]
	\begin{center}
		\includegraphics[width=0.65\linewidth]{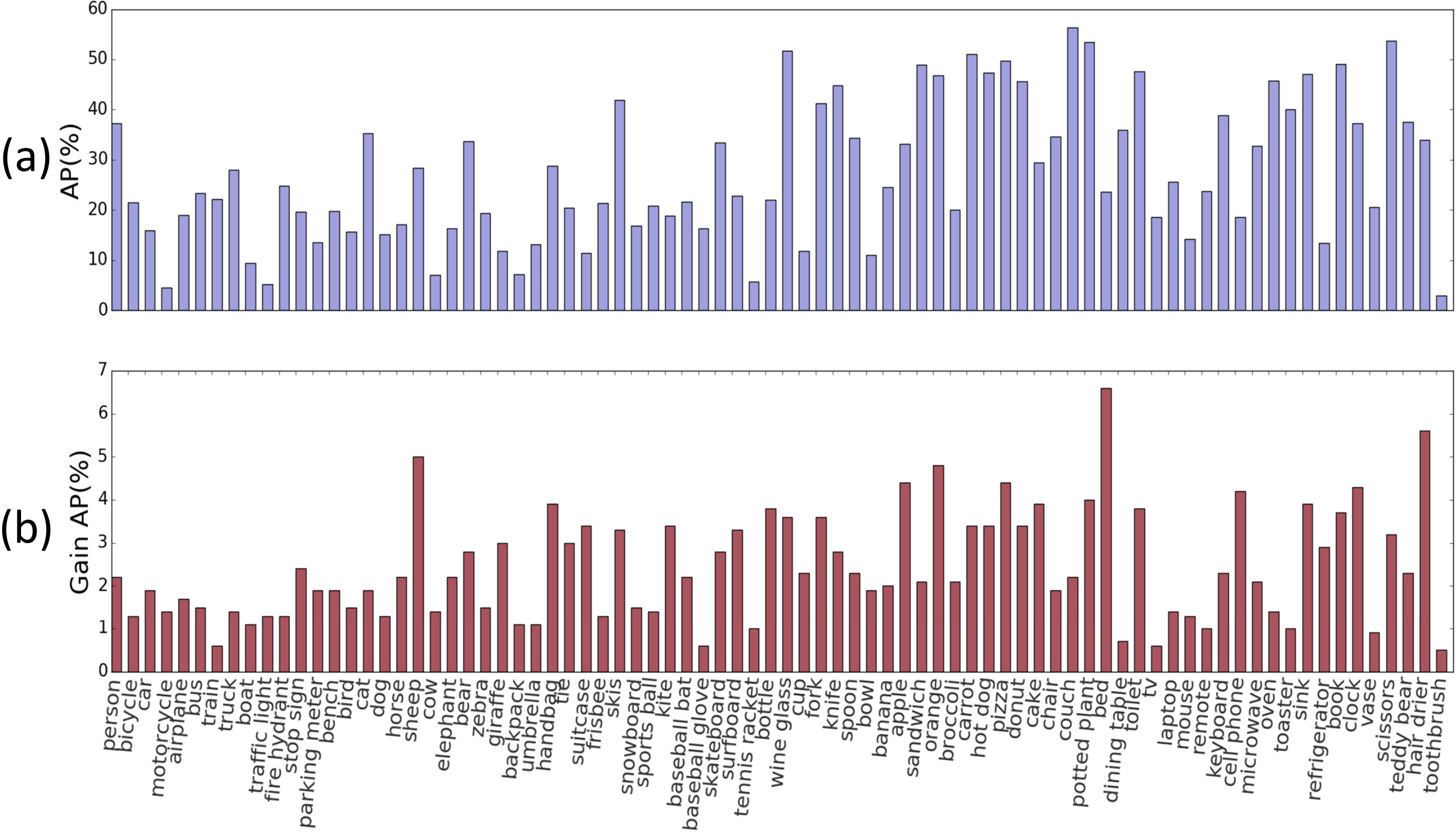}\\
		\caption{ Results of Faster R-CNN (ResNet-101) on MS COCO dataset. (a) shows AP of each class on $\Omega_{ori}$.; (b) gives gains of each class brought by $\Omega_{PSIS}$ compared with $\Omega_{ori}$.}\label{ap_after_gain}
	\end{center}\vspace{-.5cm}
\end{figure*}

\subsection{Further Analysis of Gains for Each Class}

\paragraph{Progressive Instance-Switching.} Our progressive instance-switching mainly focuses on the classes with lowest APs. To further verify its effect, we conduct experiments using Faster R-CNN (ResNet-101) on MS COCO dataset, where the settings follow Section 4.1.  Figure~\ref{pis} (a) and (b) show AP and gain ratios of each class before and after progressive instance-switching, respectively.  The gain ratios are computed by (AP$_f-$AP$_b$)/AP$_b$, where AP$_b$ and AP$_f$ are  AP of each class before and after progressive instance-switching. The results clearly demonstrate our progressive instance-switching brings much larger gain ratios for the classes with lowest APs. Meanwhile, it does not bring the side effect on the classes with highest APs, and even slightly improves their performance.

\paragraph{Progressive and Selective Instance-Switching.} To verify the effect of PSIS, we further analysis the gain AP for each class brought by PSIS using Faster R-CNN (ResNet-101) on MS COCO dataset. Figure~\ref{ap_after_gain} (a) and (b) show AP of each class on $\Omega_{ori}$ (27.3\%) and gains of each class brought by $\Omega_{PSIS}$ (29.7\%), respectively. The gains are computed by AP$_{PSIS}-$AP$_{ori}$, where  AP$_{PSIS}$ and AP$_{ori}$ are AP of each class for $\Omega_{PSIS}$ and $\Omega_{ori}$. The results clearly indicate our PSIS brings improvement for each class.

\subsection{Effect of Backbone Models with Various Depths}

In this part, we evaluate the effect of backbone models on our PSIS method. To this end, we conduct experiments using Faster R-CNN~\cite{ren2015faster} with three networks of different depths, including ResNet-50~\cite{he2016deep}, ResNet-101~\cite{he2016deep} and ResNet-152~\cite{he2016deep}. Faster R-CNN with different backbone models are trained on two datasets (i.e., $\Omega_{ori}$ and $\Omega_{PSIS}$), and results are reported on the validation set for comparison. We employ the same settings in Section 4.1 (ResNet-101 as backbone) to train Faster R-CNN with ResNet-50 and ResNet-152. The results are summarized in Table~\ref{deepnet}, where our $\Omega_{PSIS}$ achieves $1.5\%$, $2.4\%$ and $2.3\%$ gains in terms of $IoU=[0.5:0.95]$ over the original $\Omega_{ori}$ under ResNet-50, ResNet-101 and ResNet-152, respectively. Above results demonstrate that our PSIS can improve detection performance under backbone models with various depths. It is worth mentioning that Faster R-CNN with ResNet-50 trained on our $\Omega_{PSIS}$ is superior to one with ResNet-101 trained on $\Omega_{ori}$. Meanwhile, Faster R-CNN + ResNet-101 + $\Omega_{PSIS}$ outperforms  Faster R-CNN + ResNet-152 + $\Omega_{ori}$. They show again the effectiveness of our PSIS under backbone models with various depths.

\subsection{Effects of Training Settings on BlitzNet and SNIPER}

\paragraph{Half of Batch-size and 2 Times Epochs.} Due to the limited computing resources, we use half of batch-size and double training epochs in the original settings for training both BlitzNet~\cite{dvornik2017blitznet} and SNIPER~\cite{singh2018sniper}. Here we evaluate the effect of such training settings on BlitzNet and SNIPER. To this end, we use our training settings (i.e., half of batch-size and double training epochs) to re-train BlitzNet and SNIPER on the original training sets of MS COCO 2014 and MS COCO 2017, respectively. The corresponding method is indicted by $\Omega_{ori}^{\dagger}$. As compared in Tables~\ref{sniper} and~\ref{BlitzNet1}, we can see that $\Omega_{ori}^{\dagger}$ and $\Omega_{ori}$ achieve very comparable results under all evaluation metrics, clearly showing our modified training settings have little effect on performance of BlitzNet and SNIPER. Therefore, we can owe the improvement of $\Omega_{PSIS}$ over $\Omega_{ori}$ to the effectiveness of our PSIS method.

\paragraph{Negative Chip Mining for SNIPER.} Besides, we implement SNIPER~\cite{singh2018sniper} on our $\Omega_{PSIS}$ with negative chip mining, which is very time-consuming due to our limited computing resources. As listed in Table~\ref{sniper}, SNIPER with our training settings (i.e., half of batch-size and double training epochs) trained on the original dataset $\Omega_{ori}$ achieves very comparable results with the ones reported in the original paper~\cite{singh2018sniper}. When negative chip mining is employed, SNIPER trained on our $\Omega_{PSIS}$ respectively achieves $0.7\%$ and $0.6\%$ gain over one with $\Omega_{ori}^{\dagger}$ and $\Omega_{ori}$, achieving state-of-the-art performance on MS COCO 2017 dataset. Above results demonstrate our proposed PSIS is also complementary to negative chip mining strategy.

\begin{table}[!h]
	\begin{center}
		\footnotesize
		\caption{Results (\%) of BlitzNet (ResNet-50)  with different training configurations on MS COCO 2014. The results of $\Omega_{ori}$ are copied from \cite{dvornik2017blitznet}.}\label{BlitzNet1}	
		\begin{tabular}{l|ccc|ccc}
			\hline
			\multirow{2}{*}{Training set}  & \multicolumn{3}{c|}{Avg.Precision, IOU:} & \multicolumn{3}{c}{Avg.Precision, Area:} \\
			\cline{2-7}
			& 0.5:0.95 & 0.50 & 0.75 &  Small &  Med. & Large \\
			\hline
			$\Omega_{ori}^{\dagger}$  & 27.2 & 45.8 & 27.8 & 10.2 & 26.9 &  44.9\\
			$\Omega_{ori}$  & 27.3 & 46.0 & 28.1 & 10.7 & 26.8 & 46.0\\
			$\Omega_{PSIS}$  & \textbf{30.8} & \textbf{50.0} & \textbf{32.2} & \textbf{12.6} & \textbf{31.0} & \textbf{50.2} \\
			\hline
		\end{tabular}
	\end{center}
\end{table}

\begin{figure*}[!h]
	\begin{center}
		\includegraphics[width=0.85\linewidth]{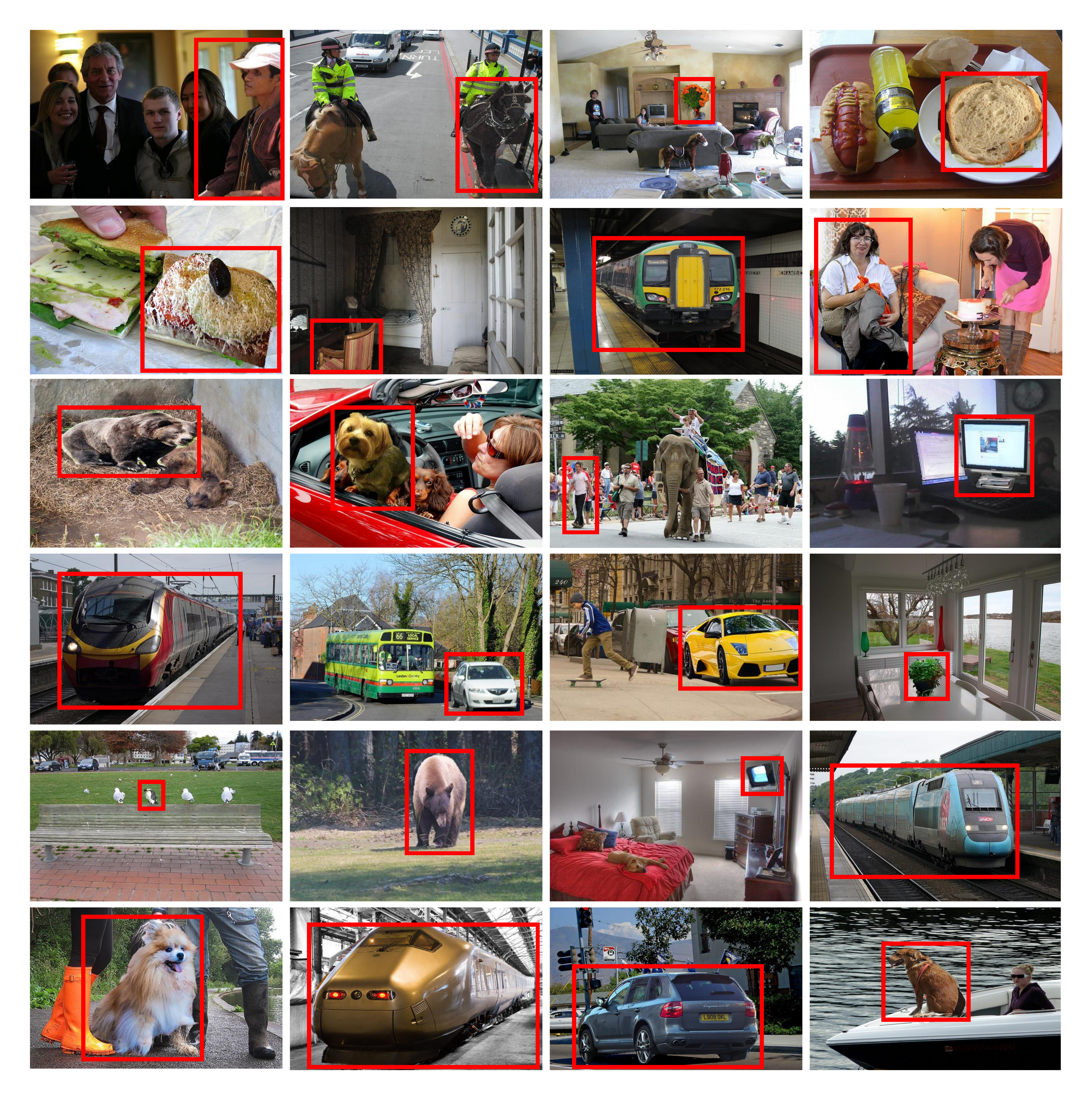}\\
		\caption{Some examples of synthetic images generated by our IS. The new instances are denoted in red boxes.}\label{good1}
	\end{center}
\end{figure*}

\end{document}